\documentclass[10pt,twocolumn,letterpaper]{article}

\usepackage[pagenumbers]{cvpr} 
\definecolor{cvprblue}{rgb}{0.21,0.49,0.74}
\usepackage[pagebackref,breaklinks,colorlinks,allcolors=cvprblue]{hyperref}

\usepackage[capitalize]{cleveref}

\newcommand{\ourmodel}{\textit{AD-R1}}

\usepackage{tabularx}
\usepackage{colortbl}
\usepackage{caption}
\usepackage{makecell}
\definecolor{mygray}{gray}{.95}
\definecolor{myblue}{HTML}{ECF2F8}
\usepackage{graphicx}
\definecolor{cvprblueplusplus}{rgb}{0.1, 0.3, 0.95}
\newcommand{\plus}[1]{\textcolor{cvprblueplusplus}{\scalebox{0.92}[0.92]{{(+#1)}}}}

\def\paperID{}
\def\confName{CVPR}
\def\confYear{2026}

\title{AD-R1: Closed-Loop Reinforcement Learning for End-to-End Autonomous Driving with Impartial World Models}

\author{Tianyi Yan$^{1,2}$, Tao Tang$^3$, Xingtai Gui$^1$, Yongkang Li$^4$, Jiasen Zheng$^5$, Weiyao Huang$^3$, \\
Lingdong Kong$^6$,  Wencheng Han$^1$, Xia Zhou$^2$, Xueyang Zhang$^2$,  Yifei Zhan$^2$, \\Kun Zhan$^2$, Cheng-zhong Xu$^1$, Jianbing Shen$^{1*}$\\
$^{1}$SKL-IOTSC, University of Macau, 
$^{2}$Li Auto Inc., 
$^{3}$Sun Yat-sen University,  \\
$^{4}$Huazhong University of Science and Technology,
$^{5}$Northwestern University, \\
$^{6}$National University of Singapore 
}

\begin{document}
\maketitle
\begin{abstract}

End-to-end models for autonomous driving hold the promise of learning complex behaviors directly from sensor data, but face critical challenges in safety and handling long-tail events.  
Reinforcement Learning (RL) offers a promising path to overcome these limitations, yet its success in autonomous driving has been elusive. We identify a fundamental flaw hindering this progress: a deep-seated ``optimistic bias" in the world models used for RL. Trained exclusively on safe expert data, these models fail to predict the consequences of erroneous actions. When conditioned on an unsafe trajectory, they hallucinate an implausibly safe future—obstacles vanish, roads are ignored—rather than forecasting the impending failure. This inability to ``imagine failure" makes them unreliable critics for learning robust driving policies. 
To address this, we introduce a framework for post-training policy refinement built around an Impartial World Model. Our primary contribution is to teach this model to be honest about danger. We achieve this with a novel data synthesis pipeline, Counterfactual Synthesis, which systematically generates a rich curriculum of plausible collisions and off-road events. This transforms the model from a passive scene completer into a veridical forecaster that remains faithful to the causal link between actions and outcomes. We then integrate this Impartial World Model into our closed-loop RL framework, where it serves as an internal critic. During refinement, the agent queries the critic to ``dream" of the outcomes for candidate actions. 
We demonstrate through extensive experiments, including on a new Risk Foreseeing Benchmark, that our model significantly outperforms baselines in predicting failures. Consequently, when used as a critic, it enables a substantial reduction in safety violations in challenging simulations, proving that teaching a model to dream of danger is a critical step towards building truly safe and intelligent autonomous agents.
\end{abstract}    
\section{Introduction}
\label{sec:intro}

\begin{figure}[tbp]
    \centering
    \includegraphics[width=\linewidth]{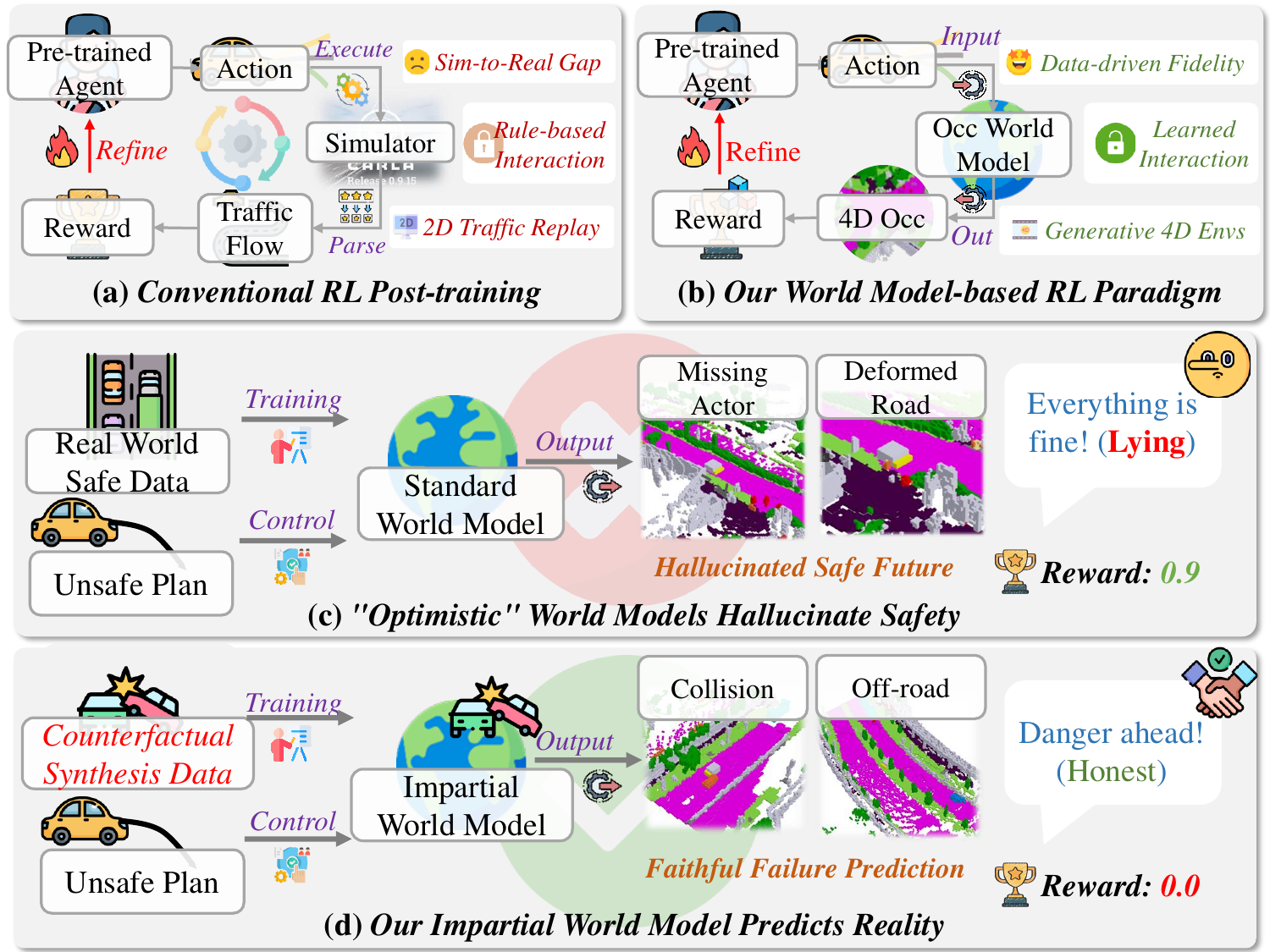}
    \vspace{-0.6cm}
    \caption{(a) \textbf{Conventional RL Post-training} relies on external simulators, suffering from a sim-to-real gap and heuristic rewards. (b) \textbf{Our Paradigm} uses a learned Occupancy World Model as an internal, generative simulator, enabling learned interactions and physically-grounded rewards. (c) \textbf{The ``Optimistic Bias"}: standard world models, however, trained only on safe data, fail to predict danger. They hallucinate safe futures for unsafe plans, providing dangerously high rewards. (d) \textbf{Our Impartial World Model}, trained with Counterfactual Synthesis Data, learns to imagine failure. It faithfully predicts the hazardous outcome, providing a correct, punitive reward, and enabling the agent to learn safely.
}
    \label{fig:teaser}
        \vspace{-0.6cm}
\end{figure}
End-to-end autonomous driving~\cite{hu2023uniad,jiang2023vad,chen2024vadv2,li2025recogdrive,wote,law,liao2025diffusiondrive,xing2025goalflow} has recently revolutionized the field, but policies trained with Imitation Learning (IL) still face significant challenges in real-world deployment, such as failures in long-tail events due to distributional shift~\cite{gao2025rad,li2025recogdrive}.
Consequently, the field is increasingly shifting towards closed-loop RL~\cite{guo2025deepseek,ouyang2022rlhf,lu2025vla-rl} through active environmental interaction to enhance decision-making capabilities of agents. 

However, the path to applying RL effectively in autonomous driving is fraught with challenges, and its widespread success has remained elusive. A fundamental obstacle stems from the very environment where RL agents are typically trained. Conventional approaches~\cite{li2024think2drive,yang2025raw2drive,li2025recogdrive} must rely on external simulators~\cite{dauner2024navsim,dosovitskiy2017carla}, which introduce their own set of limitations. 
These include the critical sim-to-real gap inherent in graphics 
engines~\cite{dosovitskiy2017carla,li2024think2drive,yang2025raw2drive}, the non-interactive nature and lack of 3D geometric awareness in traffic-flow simulators~\cite{dauner2024navsim,li2025recogdrive,liu2025reinforced}. 
These persistent issues create a major bottleneck, limiting the transferability and effectiveness of RL-trained policies in the real world.
These persistent issues necessitate a paradigm shift. 
Inspired by the remarkable success of recent generative models~\cite{gao2023magicdrive,gao2024magicdrivedit,wang2024drivedreamer,zhao2025drivedreamer2}, we pursue an alternative: learning a 3D/4D world model as a high-fidelity, generative simulator~\cite{yang2024drivearena,yan2024drivingsphere}. 
At the heart of this paradigm lies the model's ability to answer the critical question \textit{``What will happen if I take this action?"}, by generating a holistic future scene that explicitly models the dynamic interactions between the ego-vehicle, the 3D/4D environment, and other agents.

While this generative approach holds immense promise, our work identifies a systemic and perilous flaw in these state-of-the-art models, which we term \textbf{\textit{optimistic bias}} (\cref{fig:teaser}(c)).
When conditioned on an unsafe trajectory, these models refuse to predict the inevitable failure. Instead, they hallucinate an implausibly safe future. For instance, given a trajectory aimed at a pedestrian, the model will not forecast the collision; it might show the pedestrian vanishing entirely. Similarly, if commanded to drive onto a lawn, it may predict the grass magically turning into asphalt. 
This phenomenon reveals a fundamental breakdown in causal fidelity, where the model actively invents a safe outcome rather than forecasting the hazardous one commanded by the input. This makes it a dangerously unreliable critic for policy learning. We systematically quantify this bias with our proposed Risk Foreseeing Benchmark (RFB) (\cref{sec:rfb}).

To this end, we introduce a plug-and-play closed-loop RL framework, \ourmodel, to enhance the safety of any driving agent.
At the heart of our framework is an internal critic, or more precisely, a high-fidelity simulation engine, powered by a specially designed impartial world model. This engine takes the agent's planned trajectories as input, ``dreams" of their future consequences in a high-fidelity simulation, and provides dense, safety-critical feedback. This allows the agent to learn from a vast spectrum of imagined mistakes in a safe, offline setting, effectively honing its policy to navigate away from potential failures.

The success of this framework, however, is predicated on a world model that can be honest about danger. Thus, our primary technical contribution is to build an Impartial World Model (IWM) that is trained to be veridical about the outcomes of both safe and unsafe actions alike.
We build the IWM by tackling the optimistic bias at its source with \textbf{Counterfactual Synthesis}. This data pipeline treats real-world driving logs as editable stages, programmatically generating a rich curriculum of physically-consistent and causally-faithful ego-trajectories that lead to plausible failures. To further enforce causal fidelity, we introduce model-centric refinements, including a Trajectory-Aware Gating module and an Ego-Trajectory Fidelity Loss, which ensure the model's predictions remain faithful to the commanded trajectory, even when it leads to a hazardous outcome.

This resulting IWM is then integrated as the core of our RL-based critic. Its ability to realistically simulate negative outcomes unlocks a new level of granularity and physical grounding in our reward modeling. Instead of relying on simple heuristics, we can now formulate precise, multi-faceted reward signals directly from the predicted 4D occupancy. This allows us to heavily penalize not only overt failures like collisions and off-road excursions but also subtle. 
unsafe behaviors such as violating spatio-temporal buffers or ignoring vertical clearance constraints, providing a rich and reliable signal for policy refinement.

The main contributions of our work are as follows:

\begin{itemize}
    \item We introduce \ourmodel, a novel RL framework that unlocks policy refinement directly in the full 4D spatio-temporal domain. It leverages a generative occupancy world model as a high-fidelity "dream engine," enabling agents to learn safely from a vast spectrum of imagined failures—a critical capability for real-world autonomy.
    \item To enable this, we pioneer the systematic diagnosis of the \textbf{``Optimistic Bias"} in world models. We forge the \textbf{Risk Foreseeing Benchmark (RFB)}, the first benchmark designed to rigorously quantify a model's ability to predict catastrophic failure, providing a critical new tool for the community.
    
    \item Our solution is the \textbf{Impartial World Model (IWM)}, engineered to be a veridical judge of risk. We build it via two synergistic thrusts: a data-centric pipeline, \textbf{Counterfactual Synthesis}, to teach it a rich curriculum of plausible failures; and model-centric refinements \textbf{(TAG, Fidelity Loss)} to ensure its predictions remain causally faithful to the agent's commands.
    \item Extensive experiments on navsim demonstrate \ourmodel's plug-and-play effectiveness across two baselines, boosting planning PDSM by 1.7\% absolute. 
\end{itemize}

\section{Related Work}
\label{sec:formatting}
\subsection{World Models for Autonomous Driving}
The concept of a world model as a generative data engine to create high-fidelity, controllable driving videos has recently seen remarkable success~\cite{gao2023magicdrive, gao2024magicdrivedit, wang2024drivedreamer, zhao2025drivedreamer2,yan2025rlgf,liang2025worldlens,tang2025omnigen}. Frameworks like DriveArena~\cite{yang2024drivearena} and DrivingSphere~\cite{yan2024drivingsphere} have pushed the boundaries of generative closed-loop simulation. Beyond general video synthesis, a parallel and crucial application of world models lies in their use as explicit, action-conditioned simulators for policy learning and planning. Approaches have explored everything from future video prediction~\cite{gao2024vista, jia2023adriver, zhang2025epona} to, more recently, explicit 3D/4D representations~\cite{kong20253d, yan2025olidm}. 
Voxel-based world models, in particular, have demonstrated remarkable capabilities in trajectory-conditioned future state prediction, capturing fine-grained 3D geometry and 4D spatio-temporal dynamics~\cite{bian2024dynamiccity, gu2024dome, shi2025come, zheng2024occworld,wang2024occsora,liao2025i2,deng2025best3dscenerepresentation,deng2025mcnslammultiagentcollaborativeneural,yan2022plug}. 
A parallel line of work leverages world models not as explicit simulators, but to enhance policy learning through auxiliary tasks, such as learning a latent scene representation~\cite{law, zheng2025world4drive} or predicting future BEV features~\cite{wote}. 
While this ensures fidelity for on-distribution scenarios, we argue that it inherently creates the "optimistic bias"—a systemic failure to forecast the consequences of unsafe actions. This data-induced limitation makes them unreliable for safety-critical tasks like RL-based policy refinement. Our work is among the first to systematically diagnose and address this fundamental flaw, transforming the world model from a passive scene predictor into an active and \textbf{impartial} judge of safety by explicitly training it on a curriculum of synthesized failures.



\subsection{End-to-end Autonomous Driving}

End-to-end autonomous driving aims to learn a single unified policy from raw sensors to control. While modern architectures based on Transformers~\cite{hu2023uniad, jiang2023vad,chen2024vadv2,chitta2022transfuser,li2024hydra} and Diffusion Models~\cite{liao2025diffusiondrive, xing2025goalflow} have achieved significant success, they largely follow an Imitation Learning (IL) paradigm. A fundamental limitation of pure IL is its reliance on the expert data distribution; models struggle with causal confusion and are prone to failures in long-tail scenarios due to distributional shift.
To overcome these challenges, a growing consensus is forming around the need to incorporate closed-loop Reinforcement Learning (RL). 
Early efforts explored model-free RL in simulators like CARLA~\cite{dosovitskiy2017carla, renz2024carllava}, but sample inefficiency remains a major hurdle for real-world application. 
Consequently, model-based RL has become a more promising direction. Approaches like Think2Drive~\cite{li2024think2drive} and Raw2Drive~\cite{yang2025raw2drive} learn a world model within the CARLA simulator, but this paradigm inherits the simulator's own sim-to-real gap. Some efforts used simulators with non-reactive 2D layouts~\cite{li2025recogdrive}, their inability to model complex 3D/4D interactions is a major drawback. A significant breakthrough came with reconstruction-based simulation, such as RAD~\cite{gao2025rad}, which leverages 3D Gaussian Splatting for photo-realism. 

Instead of reconstructing a single, static scene, we utilize a learned world model to \textit{generate} a vast diversity of potential future outcomes in response to the agent's actions. Our primary contribution is an impartial world model that acts as this reliable generative simulator, or \textit{internal critic}. This enables a generative world model-based RL framework where the agent can effectively and safely learn from imagined mistakes, leading to a demonstrably safer driving policy.

\section{Methodology}
\label{sec:preliminaries}

\begin{figure*}[tbp]
    \centering
    \includegraphics[width=\linewidth]{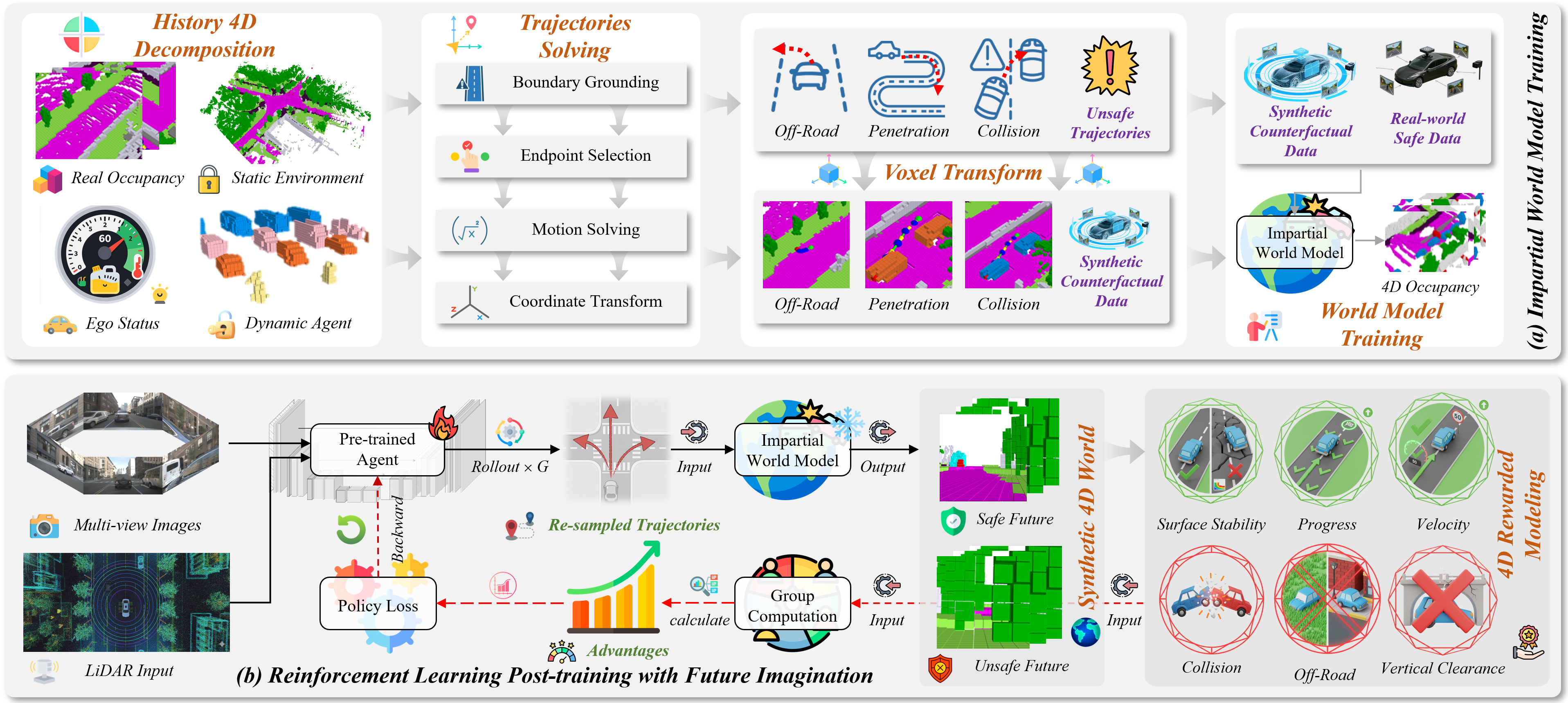}
    \vspace{-0.6cm}
    \caption{\textbf{Overview of \ourmodel.} Our framework involves two stages: (a) Impartial World Model Training: We first introduce a Counterfactual Synthesis pipeline that decomposes real scenes and programmatically generates unsafe trajectories to create Synthetic Negative Data. Our Impartial World Model is then trained on a mix of this synthetic failure data and real-world safe data. (b) Reinforcement Learning Post-training: The trained model acts as an internal critic. It takes candidate trajectories from a pre-trained agent, "dreams" of the 4D future outcomes, and our 4D Rewarded Modeling module computes a dense reward based on the imagined world. This provides a strong Policy Loss to refine the agent's safety and robustness through imagined failures.}
    \label{fig:overview}
\end{figure*}

\subsection{Overview and Problem Formulation}



Our goal is to refine a pre-trained end-to-end driving policy, $\pi$, which maps sensor observations $\mathcal{O}_t$ and a command $\mathcal{G}$ to a future ego-trajectory $\mathcal{T}_{\text{ego}}$. While often trained with imitation learning, such policies struggle in long-tail events~\cite{li2025recogdrive,gao2025rad}, we formulate this as a reinforcement learning (RL) problem, where the policy $\pi$ is optimized to maximize the expected future reward from a reward function $R$:
\begin{equation}
    \pi^* = \arg\max_{\pi} \mathbb{E}_{\mathcal{T}_{\text{ego}} \sim \pi(\mathcal{O}_t, \mathcal{G})} \left[ R(\mathcal{T}_{\text{ego}}, \mathcal{O}_t) \right],
    \label{eq:rl_objective}
\end{equation}
where the reward function $R$ evaluates the quality of a proposed trajectory $\mathcal{T}_{\text{ego}}$. 

To obtain this reward signal safely and efficiently, existing methods~\cite{dauner2024navsim,li2025recogdrive,gao2025rad} rely on external simulators. However, as discussed, these are limited by sim-to-real gaps in graphics engines, the non-interactive nature of 2D layout simulators, or the static, costly nature of reconstruction-based environments. 
In a departure from these paradigms, we introduce a learned world model, $\mathcal{M}$, to serve as a high-fidelity, internal generative simulator.
$\mathcal{M}$ takes the historical observations $\mathcal{O}_{\le t}$ and a candidate ego-trajectory $\mathcal{T}_{\text{ego}}$ as input, and predicts the corresponding future evolution of the scene, typically represented as a sequence of future 3D occupancy grids $\{\hat{\mathcal{O}}_{t+k}\}_{k=1}^K$:
\begin{equation}
    \{\hat{\mathcal{O}}_{t+k}\}_{k=1}^K = \mathcal{M}(\mathcal{O}_{\le t}, \mathcal{T}_{\text{ego}}).
    \label{eq:world_model_definition}
\end{equation}

The reward $R$ is then computed directly from this predicted future. Therefore, the entire framework hinges on the model's fidelity. 
A good world model must faithfully predict the consequences of any given trajectory, especially unsafe ones, to provide a reliable reward signal. 

As shown in \cref{fig:overview}, our method is thus composed of two key components: (1) the design of a novel \textbf{Impartial World Model} (detailed in Sec.~\ref{sec:fawm}), and (2) its integration into our RL framework for policy refinement (detailed in Sec.~\ref{sec:rl}).

\subsection{Impartial Occupancy World Model}
\label{sec:fawm}
While state-of-the-art world models exhibit impressive forecasting performance, they suffer from the ``optimistic bias" discussed previously. To address this, we build an occupancy-based prediction model designed to capture the complex 3D geometry and 4D spatio-temporal dynamics of the driving scene. Crucially, this model is fundamentally engineered to overcome this bias. We first outline the backbone architecture before detailing our synergistic data-centric and model-centric refinements.

\subsubsection{Backbone Architecture.}
Our impartial world model builds upon a powerful and efficient architecture inspired by I²-World~\cite{liao2025i2world}. The model operates in a latent space for computational efficiency, following a two-stage process:
\begin{itemize}
    \item \textbf{Scene Tokenization:} An encoder, termed the I²-Scene Tokenizer, first compresses the sequence of high-dimensional historical occupancy grids $\mathcal{O}_{\le t}$ into a compact sequence of latent tokens $\mathcal{B}_{\le t}$. This tokenizer is designed to efficiently capture both the static spatial details within each frame (intra-scene) and the dynamic temporal evolution across frames (inter-scene).
    \item \textbf{Autoregressive Forecasting:} A transformer-based forecaster, the I²-Former, then autoregressively predicts a sequence of future latent tokens $\{\hat{\mathcal{B}}_{t+k}\}_{k=1}^K$. This forecaster consists of an \textbf{encoder} that processes the historical tokens and the candidate ego-trajectory, and a \textbf{decoder} that generates the future tokens. The final sequence of occupancy grids $\{\hat{\mathcal{O}}_{t+k}\}_{k=1}^K$ is produced by decoding these latent predictions.
\end{itemize}
This robust two-stage architecture provides the foundation upon which we introduce our key refinements to instill the vital capability of imagining failure.

\subsubsection{Counterfactual Data Synthesis.}

Existing real-world datasets~\cite{caesar2020nuscenes, sun2020waymo}, which form the foundation of most world models, primarily consist of professional trajectories recorded in safe driving scenarios.
The inherent lack of failure examples in this data is the primary cause of optimistic bias. We address this at its source by proposing a systematic data synthesis pipeline to create a diverse and structured curriculum of counterfactual negative samples.

Our pipeline treats each safe scenario from the dataset as an editable stage. Given the historical occupancy sequence $O_{\le t}$, we first decompose the current frame $O_t$. This decomposition leverages the ground-truth 3D annotations provided by the nuScenes dataset~\cite{caesar2020nuscenes,tian2023occ3d} to isolate the static environment, dynamic agents, and the original ego-vehicle's state using clustering~\cite{schubert2017dbscan}. This provides the necessary context to generate plausible, scene-aware failure trajectories. We acknowledge that this reliance on annotations is a current limitation, though we believe it could be replaced by a high-fidelity perception system in the future.

Next, we programmatically generate a new, unsafe ego-trajectory $\tilde{T}_{\text{ego}} = \{\mathbf{p}_{t+k}\}_{k=1}^K$. All unsafe trajectories are generated by a \textbf{unified kinematic model}. This model steers the ego-vehicle from its current state toward a designated \textbf{target point} $\mathbf{p}_{\text{target}}$, which is selected based on the desired failure mode. The future position $\mathbf{p}_{t+k}$ at each timestep $k$ is determined by blending the vehicle's historical momentum with a vector towards the target:
\begin{equation}
    \mathbf{v}_{\text{target}} = \frac{\mathbf{p}_{\text{target}} - \mathbf{p}_{t}}{\|\mathbf{p}_{\text{target}} - \mathbf{p}_{t}\|}
\end{equation}
\begin{equation}
    \mathbf{p}_{t+k} = \mathbf{p}_{t} + \alpha_k \cdot \Delta d_k \cdot \mathbf{h}_{\text{hist}} + \beta_k \cdot \Delta d_k \cdot \mathbf{v}_{\text{target}} + \mathcal{N}(0, \sigma^2)
    \label{eq:kinematic_model}
\end{equation}
where $\mathbf{h}_{\text{hist}}$ is the historical heading, $\mathbf{v}_{\text{target}}$ is the unit vector towards the target, $\Delta d_k$ is the step distance, and $\alpha_k, \beta_k$ are time-varying blending coefficients. These coefficients are designed to create a smooth, plausible maneuver that starts from the current trajectory and progressively deviates towards the target, mimicking a realistic driving action.

We construct a hierarchical curriculum of failures by defining strategies for selecting the target point $\mathbf{p}_{\text{target}}$:

\noindent\textbf{Ego-State Failure (Off-Road Events).} To simulate leaving the roadway, $\mathbf{p}_{\text{target}}$ is selected as a point just outside the nearest drivable area boundary. Applying our kinematic model (Eq. \ref{eq:kinematic_model}) generates a smooth off-road drift.

\noindent\textbf{Interaction with Static Environment (Penetration).} The next level of complexity involves interacting with non-movable parts of the world. We generate trajectories that lead to the ego-vehicle penetrating static obstacles such as buildings, barriers, or vegetation. This is achieved by planning a path directly into a nearby static object identified from the decomposed scene map. This teaches the model the concept of solid objects and physical impassability.

\noindent\textbf{Interaction with Dynamic Agents (Collisions).} To simulate a dynamic collision with another agent, $\mathbf{p}_{\text{target}}$ is set to a future predicted position.
For simplicity, we assume a linear or constant-velocity prediction for the target agent, and our kinematic model generates an intercept trajectory.

By using a single, flexible trajectory generation model applied to a structured hierarchy of targets, we can efficiently synthesize a rich and diverse curriculum. 
Finally, we synthesize the new ground-truth future occupancy sequence $\{\mathcal{O}'_{t+k}\}_{k=1}^K$. For each future timestep $k$, we compute the rigid-body transformation $\mathbf{T}_{k}$ that maps voxel coordinates from the new counterfactual ego-frame back to the original one. The new future grid $\mathcal{O}'_{t+k}$ is then generated by applying this transformation to the original grid $\mathcal{O}_{t+k}$, using interpolation for non-integer coordinates:
\begin{equation}
\label{eq:synthesis_transform}
\mathcal{O}'_{t+k}(\mathbf{v}') = \mathcal{O}_{t+k}(\mathbf{T}_{k}(\mathbf{v}')).
\end{equation}
This operation, using interpolation for non-integer coordinates, ensures that all complex spatial relationships are preserved with high physical fidelity. 
By training on this rich curriculum of synthesized negative samples, we enable our world model $\mathcal{M}$ to learn the crucial causal link between unsafe ego-trajectories and their corresponding failure outcomes, such as collisions and off-road events.
\subsubsection{Model-Centric Refinements}
\noindent\textbf{Trajectory-Aware Gating (TAG).}
To further enhance the model's fidelity to control input, we introduce a lightweight, plug-and-play gating module. 
Specifically, we first rasterize the input ego-trajectory $\mathcal{T}_{\text{ego}}$ into a 2D spatial mask, which a small CNN processes into a gating map, $\mathcal{A}_{\text{traj}} \in [0, 1]^{h \times w}$.
This map dynamically modulates the intermediate scene features $\mathcal{F}_{\text{scene}}$ via element-wise multiplication:
\begin{equation}
    \mathcal{F}_{\text{gated}} = \mathcal{F}_{\text{scene}} \odot \mathcal{A}_{\text{traj}}.
\end{equation}
This forces the model to prioritize features along the commanded path, preventing it from ``ignoring" an obstacle to invent a safe outcome.

\noindent\textbf{Ego-Trajectory Fidelity Loss.}
The final piece is a novel loss function that penalizes any deviation of the predicted ego-vehicle from its commanded trajectory.
For each predicted future timestep $k$, we decode the predicted occupancy map $\hat{\mathcal{O}}_{t+k}$ and compute the geometric center of all voxels classified as the ego-vehicle, denoted as $\hat{p}_{\text{ego}}^{t+k}$. 
We then introduce an Ego-Trajectory Fidelity Loss, $\mathcal{L}_{\text{fidelity}}$, which measures the L2 distance between this predicted center and the ground-truth position from the input control trajectory, $p_{\text{ego}}^{t+k}$:
\begin{equation}
    \mathcal{L}_{\text{fidelity}} = \sum_{k=1}^{K} w_k \left\| \hat{p}_{\text{ego}}^{t+k} - p_{\text{ego}}^{t+k} \right\|_2^2.
\end{equation}
This loss acts as a strong regularizer, directly discouraging the model from hallucinating a future where the ego-vehicle autonomously ``swerves" to avoid a collision. It enforces the principle that the world model's role is to predict, not to act.

\noindent\textbf{Final Training Objective.}
Our impartial world model is trained on a mixed dataset of real and synthesized counterfactual data by augmenting the original baseline objective $\mathcal{L}_{\text{baseline}}$~\cite{liao2025i2} with our proposed fidelity loss:
\begin{equation}
    \mathcal{L}_{\text{total}} = \mathcal{L}_{\text{baseline}} + \lambda \mathcal{L}_{\text{fidelity}},
\end{equation}
where $\lambda$ is a balancing hyperparameter. The model is trained on a mixed dataset comprising both original safe scenarios and our synthesized failure scenarios.

\subsection{RL with Future Imagination}
\label{sec:rl}
\subsubsection{Training Paradigm}
Our training paradigm leverages our impartial world model as a high-fidelity simulator to refine a pre-trained policy $\pi$ in a closed-loop, yet entirely offline, manner. To enable stable and efficient exploration beyond simple imitation, we introduce a Group Relative Policy Optimization (GRPO) algorithm~\cite{guo2025deepseek}, adapted for trajectory refinement.

Concretely, for a given observation $\mathcal{O}_t$, we refine a diffusion-based driving policy $\pi_\theta$.
Following~\cite{li2025recogdrive}, we formulate the diffusion denoising process as a finite-horizon Markov Decision Process (MDP), where the diffusion planner acts as a stochastic policy.
Starting from Gaussian noise $\mathcal{T}_T \sim \mathcal{N}(0, \mathbf{I})$, the policy gradually denoises over $T$ steps to generate a full trajectory $\mathcal{T}_{\text{ego}} = \mathcal{T}_0$.
To enable exploration beyond imitation, we sample a group of $G$ trajectory rollouts, $\{\mathcal{T}_{\text{ego}}^{(i)}\}_{i=1}^G$.
Each trajectory is then passed to our world model $\mathcal{M}$ to ``dream'' of its consequences, yielding a corresponding future occupancy sequence $\hat{S}^{(i)} = \mathcal{M}(\mathcal{O}_t, \mathcal{T}_{\text{ego}}^{(i)})$. 
We then compute a scalar reward $r_i = \mathcal{R}(\hat{S}^{(i)})$ (\cref{sec:reward}) for each imagined rollout using our reward function.

Unlike prior approaches that rely on simple surrogate rewards, our physically-grounded rewards provide realistic feedback on safety and comfort. We treat the entire group of rollouts as a single-step decision process. To obtain a stable learning signal, we compute a group-standardized advantage for each trajectory:
\begin{equation}
    \hat{A}_i = \frac{r_i - \text{mean}(\{r_j\}_{j=1}^G)}{\text{std}(\{r_j\}_{j=1}^G)}, \quad i=1, \dots, G.
\end{equation}
This normalization centers the rewards and scales them by their variance, focusing the policy update on the relative quality of trajectories within the sampled group.

Finally, we compute the policy loss. The objective is composed of two main terms: a reinforcement learning loss that encourages reward-seeking behavior, and a behavior cloning loss that regularizes the policy to prevent catastrophic forgetting.

The RL loss, based on the GRPO formulation, is computed over the group of sampled trajectories:
\begin{equation}
\label{eq:loss_rl}
\mathcal{L}_{\text{RL}} = - \frac{1}{G} \sum_{i=1}^{G} \log \pi_\theta(\mathcal{T}_{\text{ego}}^{(i)} \mid \mathcal{O}_t) \hat{A}_i.
\end{equation}
This term updates the policy to increase the likelihood of trajectories that yield a higher relative advantage within the group.

To prevent the policy from deviating too far from the sensible priors of the pre-trained model, we add a behavior cloning (BC) loss. This term anchors the refined policy $\pi_\theta$ to the original reference policy $\pi_{\text{ref}}$:
\begin{equation}
\label{eq:loss_bc}
\mathcal{L}_{\text{BC}} = \mathbb{E}_{\mathcal{T}_{\text{ref}} \sim \pi_{\text{ref}}} \left[ -\log \pi_\theta(\mathcal{T}_{\text{ref}} \mid \mathcal{O}_t) \right].
\end{equation}

The final training objective is a weighted sum of these two components:
\begin{equation}
\label{eq:loss_total}
\mathcal{L}_{\text{total}} = \mathcal{L}_{\text{RL}} + \lambda \mathcal{L}_{\text{BC}},
\end{equation}
where $\lambda$ is a hyperparameter that controls the strength of the regularization. Through this GRPO-based objective, the driving policy learns to generate safer and more comfortable trajectories in a purely offline setting, effectively learning from a vast spectrum of imagined failures.

\subsubsection{4D Reward Modeling}
\label{sec:reward}
The predicted 4D occupancy sequence, $\hat{S}$, provides a rich, physically-grounded canvas for reward modeling. Leveraging the high fidelity of our impartial world model, we design a comprehensive reward function, $\mathcal{R}(\hat{S})$, that translates this imagined future into a scalar reward signal. 
The explicit 3D and temporal nature of our predictions allows us to formulate rewards that capture safety, comfort, and task progress with a level of detail unattainable by heuristic-based or 2D BEV approaches.
Let $\mathcal{O}_t(\mathbf{v})$ be the set of semantic labels (\eg, $\{\texttt{Ego}, \texttt{Drivable}\}$) assigned to a voxel $\mathbf{v}$ at future timestep $t$. Our reward is a weighted sum of several components designed to encourage safe and efficient driving.

\subsubsection{Safety-Critical Penalties}
The foremost priority is to penalize any future that our model predicts as unsafe.

\noindent\textbf{Volumetric Collision Penalty ($\boldsymbol{\mathcal{R}_{\text{coll}}}$):}
A collision is the most severe failure event. We go beyond a simple binary penalty by quantifying the \textit{severity} of the predicted collision. For any predicted timestep $t$, we identify the set of voxels occupied by the ego vehicle, $\mathcal{V}_{\text{ego}}$, and any other dynamic agent $k$ (\eg, another vehicle or a pedestrian), $\mathcal{V}_{\text{agent},k}$. A collision occurs if their intersection is non-empty. We define the \textbf{Volumetric Collision IoU (VC-IoU)} to quantify the spatial overlap:
\begin{equation}
\label{eq:vciou}
\text{VC-IoU}_k = \frac{|\mathcal{V}_{\text{ego}} \cap \mathcal{V}_{\text{agent},k}|}{|\mathcal{V}_{\text{ego}} \cup \mathcal{V}_{\text{agent},k}|}.
\end{equation}
This metric naturally captures the extent of the overlap, providing a proxy for impact severity. The collision penalty then heavily penalizes any contact while scaling with this severity:
\begin{small}{
\begin{equation}
\mathcal{R}_{\text{coll}} = - \sum_{t,k} \left[ C_{\text{base}} \cdot \mathbb{I}(\text{VC-IoU}_k > 0) + C_{\text{iou}} \cdot \text{VC-IoU}_k \right],
\end{equation}
}\end{small}
where $\mathbb{I}(\cdot)$ is the indicator function, $C_{\text{base}}$ is a large constant penalty for any contact, and $C_{\text{iou}}$ weights the severity.

\noindent\textbf{Off-Road and Sidewalk Penalty ($\boldsymbol{\mathcal{R}_{\text{offroad}}}$):}
Leveraging semantic predictions, we penalize the ego vehicle for occupying non-drivable areas. We identify the set of voxels under the ego vehicle's footprint, $\mathcal{V}_{\text{footprint}}$. The penalty is proportional to the number of these voxels classified with non-permissible semantics:
\begin{small}{
\begin{align}
\mathcal{R}_{\text{offroad}} = - \sum_{t} |  \{ \mathbf{v} \in \mathcal{V}_{\text{footprint}}   \mid \texttt{Sidewalk} \nonumber \\ \in \mathcal{O}_t(\mathbf{v}) \lor \texttt{Vegetation} \in \mathcal{O}_t(\mathbf{v}) \} |.
\end{align}
} 
\end{small}

\subsubsection{3D/4D-Aware Comfort \& Rule-Adherence}
Here, we introduce the design of rewards uniquely enabled by the 3D and temporal awareness of our model, showcasing clear advantages over 2D BEV.

\noindent\textbf{Vertical Clearance Penalty ($\boldsymbol{\mathcal{R}_{\text{clearance}}}$):} Safe driving requires awareness of vertical space, to which 2D BEV representations are blind. Our 3D model prevents collisions with low-hanging structures (\eg, bridges, tunnels). We define a 3D ``safety bubble" above the ego vehicle, $\mathcal{V}_{\text{bubble}}$, and penalize its intersection with static obstacles:
\begin{small}{
\begin{equation}
\mathcal{R}_{\text{clearance}} = - \sum_{t} \left| \{ \mathbf{v} \in \mathcal{V}_{\text{bubble}} \mid \texttt{Building} \in \mathcal{O}_t(\mathbf{v}) \} \right|.
\end{equation}
}\end{small}
This provides a crucial safety constraint impossible to model from a BEV perspective alone.


\noindent\textbf{Drivable Surface Stability Reward ($\boldsymbol{\mathcal{R}_{\text{stability}}}$):} 
A region labeled ``drivable" in 2D may be a steep curb or uneven road in 3D. We reward the agent for choosing stable, flat surfaces by penalizing high variance in the Z-coordinate (height) of the drivable surface voxels directly beneath the ego vehicle's footprint:
\begin{small}{
\begin{align}
 \mathcal{R}_{\text{stability}} = &  - \sum_{t} \text{Var} \left( Z(\mathbf{v}) \right) \quad  
\text{for } \nonumber \\ \mathbf{v} \in \{ \text{voxels below } &  \mathcal{V}_{\text{footprint}} \mid \texttt{Drivable} \in \mathcal{O}_t(\mathbf{v}) \}.
\end{align}
}\end{small}
This encourages a smoother and safer ride.

\subsubsection{Task-Oriented Rewards}
Finally, to ensure the agent makes progress, we include standard task-oriented rewards.

\noindent \textbf{Progress Reward ($\boldsymbol{\mathcal{R}_{\text{progress}}}$):} Reward for advancing along a pre-defined route or towards a goal.

\noindent \textbf{Velocity Reward ($\boldsymbol{\mathcal{R}_{\text{velocity}}}$):} Reward for maintaining a target speed, typically the local speed limit.

The final reward passed to the RL algorithm is the weighted sum of all components:
\begin{align}
\mathcal{R}_{\text{total}} = w_{\text{coll}}\mathcal{R}_{\text{coll}} + w_{\text{offroad}}\mathcal{R}_{\text{offroad}} + w_{\text{clear}}\mathcal{R}_{\text{clearance}}  \nonumber \\ 
+ \dots + w_{\text{prog}}\mathcal{R}_{\text{progress}}.
\end{align}
By combining these multi-faceted reward signals, our framework fully exploits the rich, predictive power of the impartial 4D world model, enabling the learning of a driving policy that is not only effective but also demonstrably safer and more physically-grounded.
\section{Experiments}

\subsection{Experimental Setup}

Our impartial world model is trained on a mixed dataset. The primary source of safe driving scenarios is the nuScenes dataset~\cite{caesar2020nuscenes}, providing rich sensor data and annotations. This is augmented with failure scenarios generated via our Counterfactual Synthesis pipeline. \textbf{We use a data split of 80\% real nuScenes data and 20\% our synthesized counterfactual data. This ratio was determined through empirical validation to provide a sufficient number of failure examples without significantly shifting the model's prior away from realistic, safe driving dynamics.}
For the closed-loop evaluation of the final driving policy, we use the official NavSim benchmark~\cite{dauner2024navsim}, which offers a standardized environment.
We evaluate our method at two levels: the fidelity of the world model and the performance of the refined driving policy. 
\textbf{Detailed Metric and Implementation are included in the supplementary}.

\subsection{Quantitative Analysis}

\noindent\textbf{\ourmodel~Excels at Foreseeing Risk.}
\label{sec:rfb}
As shown in \cref{tab:rfb_results}, standard world models struggle significantly on our RFB benchmark. DOME~\cite{gu2024dome} and $I^2-$world~\cite{liao2025i2}, trained only on safe data, exhibit a strong optimistic bias, resulting in very low $f-$IoU and DAF. In contrast, our IWM, trained with our synthetic counterfactual data and model-centric refinements, achieves dramatically higher scores. This quantitatively validates our primary claim: our impartial world model can faithfully ``imagine failure" where other models hallucinate safety.

\begin{table}[t]
\centering
\caption{\textbf{World Model Performance on the Risk Foreseeing Benchmark (RFB).} \ourmodel~significantly outperforms the baseline in predicting the outcomes of unsafe trajectories, demonstrating its ability to overcome optimistic bias.}
\label{tab:rfb_results}
\vspace{-0.1cm}
\addtolength{\tabcolsep}{2.3pt}
\begin{tabularx}{\linewidth}{l|ccc}
\toprule
World Model & G-IoU (\%) $\uparrow$ & \textit{f}-IoU $\uparrow$ & DAF $\uparrow$ \\
\midrule
DOME~\cite{gu2024dome} & 21.01 & 14.21 & 8.75
\\
I²-World~\cite{liao2025i2world} & 30.91 & 21.44 & 14.21
\\
\rowcolor{mygray} \textbf{IWM~(Ours)} & \textbf{40.21} & \textbf{45.91} & \textbf{25.10} 
\\
\bottomrule
\end{tabularx}
\vspace{-0.2cm}
\end{table}

\begin{table*}[h]
\centering
\caption{\textbf{Closed-Loop Performance on NavSim.} Our post-training refinement framework, powered by \ourmodel, significantly improves the safety and planning quality of two state-of-the-art baseline agents.}
\label{tab:main_results}
\vspace{-0.1cm}
\addtolength{\tabcolsep}{4.2pt}
\begin{tabularx}{\linewidth}{lcccccccc}
\toprule
Method & Scheme & WM  & NC $\uparrow$ & DAC $\uparrow$ & TTC $\uparrow$ & Comf. $\uparrow$& EP $\uparrow$& PDMS $\uparrow$ \\
\midrule
Constant Velocity & &   &68.0&57.8& 50.0 & 100 & 19.4 & 20.6 \\
Ego Status MLP & &   &93.0&77.3& 83.6 & 100 & 62.8 & 65.6 \\
\midrule
VADv2~\cite{chen2024vadv2} & IL &   &97.2&89.1& 91.6 & 100 & 76.0 & 80.9 \\
DrivingGPT~\cite{chen2025drivinggpt} & IL &   &98.9&90.7& 94.9 & 100 & 79.7 & 82.4 \\
UniAD~\cite{hu2023uniad} & IL &   &97.8&91.9& 92.9 & 100 & 78.8 & 83.4 \\
TransFuser~\cite{chitta2022transfuser} & IL &   &97.7&92.8& 92.8 & 100 & 79.2 & 84.0 \\
Hydra-MDP~\cite{li2024hydra} & IL &   &98.3&96.0& 94.6 & 100 & 78.7 & 86.5 \\
LAW~\cite{law} & IL+SSL & \checkmark  &97.2&93.3& 91.9 & 100 & 78.8 & 83.8 \\
World4Drive~\cite{zheng2025world4drive} & IL & \checkmark  &97.4&94.3& 92.8 & 100 & 79.9 & 85.1 \\
Trajhf~\cite{li2025trajhf} & IL+RL & \checkmark  &96.6&96.6& 92.1 & 100 & 84.5 & 87.6 \\
WoTE~\cite{wote} & IL & \checkmark  &98.5&96.8& 94.9 & 99.9 & 81.9 & 88.3 \\
DriveDPO~\cite{shang2025drivedpo} & IL+RL &   &98.5&98.1& 94.8 & 99.9 & 84.3 & 90.0 \\
GoalFlow~\cite{xing2025goalflow} & IL &   &98.4&98.3& 94.6 & 100 & 85.0 & 90.3 \\
Recogdrive~\cite{li2025recogdrive} & IL+RL &   &97.9&97.3& 94.9 & 100 & 87.3 & 90.8 \\
\midrule
DiffusionDrive~\cite{liao2025diffusiondrive} & IL &   &98.2&96.2& 94.7 & 100 & 82.2 & 88.1 \\
\rowcolor{mygray} + \textbf{\ourmodel~(Ours)}  & RL & \checkmark   &98.5&97.5& 94.6 & 100 & 84.8 & 89.8\plus{1.7} \\

\midrule
Recogdrive-stage2~\cite{li2025recogdrive} & IL &     &98.1&94.7& 94.2 & 100 & 80.9 & 86.5 \\
\rowcolor{mygray}  + \textbf{\ourmodel~(Ours)} & RL & \checkmark  &98.74&97.8& 94.8 & 100 & 87.5 & \textbf{91.9}\plus{1.1} \\
\bottomrule
\end{tabularx}
\end{table*}

\noindent\textbf{Post-Training Refinement Boosts Safety and Performance.}
\cref{tab:main_results} presents the core results of our paper. Applying our plug-and-play refinement framework yields substantial improvements across both baselines. For both diffusion-policy based method, DiffusionDrive~\cite{liao2025diffusiondrive} and VLA-based method Recogdrive~\cite{li2025recogdrive}, we observe a stable improvement of PDMS scores in closed-loop simulations. This demonstrates that learning from imagined failures directly translates to safer real-world behavior. Furthermore, this enhanced safety does not come at the cost of performance. Our method improves the PDSM score for both agents, with an absolute boost of 1.7\% and 1.1\%, respectively. 

\subsection{Ablation Studies}

\noindent\textbf{Effect of Core Components.}
We conduct a comprehensive ablation study to dissect the contribution of each of our proposed components, with results presented in \cref{tab:ablation}. 
We begin with a strong IL-trained baseline, DiffusionDrive. Our first key finding is that naively applying a standard, optimistic world model (\textit{e.g.}, I²-World) for RL-based refinement is actively harmful, causing a significant drop in the PDMS score (88.1 $\rightarrow$ 85.3). 

In contrast, incrementally adding our contributions demonstrates a clear and consistent path to improvement. Introducing \textbf{Counterfactual Synthesis (CS)} provides the most substantial boost, lifting the PDMS score to 89.1. This highlights that teaching the model to recognize failure is the single most critical factor. Subsequently, adding our \textbf{Model-centric Refinements (MR)} (TAG and Fidelity Loss) further improves the score to 89.3 by enforcing stricter causal fidelity. Finally, leveraging our physically-grounded \textbf{4D Rewards (R)} in the refinement loop yields the best overall performance (89.8 PDMS). This validates that each of our contributions—curing the optimistic bias, enforcing causal fidelity, and utilizing detailed 3D/4D rewards—plays a crucial and synergistic role in successfully and safely refining the driving policy.

\begin{table}[t]
\centering
\caption{\textbf{Ablation study on the components of \ourmodel~and our RL framework.} Results are reported on the G-IoU of RFB (for the world model) and NavSim PDMS (for the final policy).}
\label{tab:ablation}
\addtolength{\tabcolsep}{4.5pt}
\begin{tabularx}{\linewidth}{lcc}

\toprule
Configuration &G-IoU $\uparrow$ & PDMS $\uparrow$ \\
\midrule
Baseline (DiffusionDrive)  & -  & 88.1\\
+ $i^2$-world & 30.91 & 85.3\\
+ CS & 37.22 & 89.1\\
+ CS + MR  & 40.21  &89.3\\
\rowcolor{mygray} + CS + MR + R & \textbf{40.21} &\textbf{89.8} \\
\bottomrule
\vspace{-0.3cm}

\end{tabularx}
\end{table}

\subsection{Qualitative Analysis}

\begin{figure}[h]
\centering
\includegraphics[width=\linewidth]{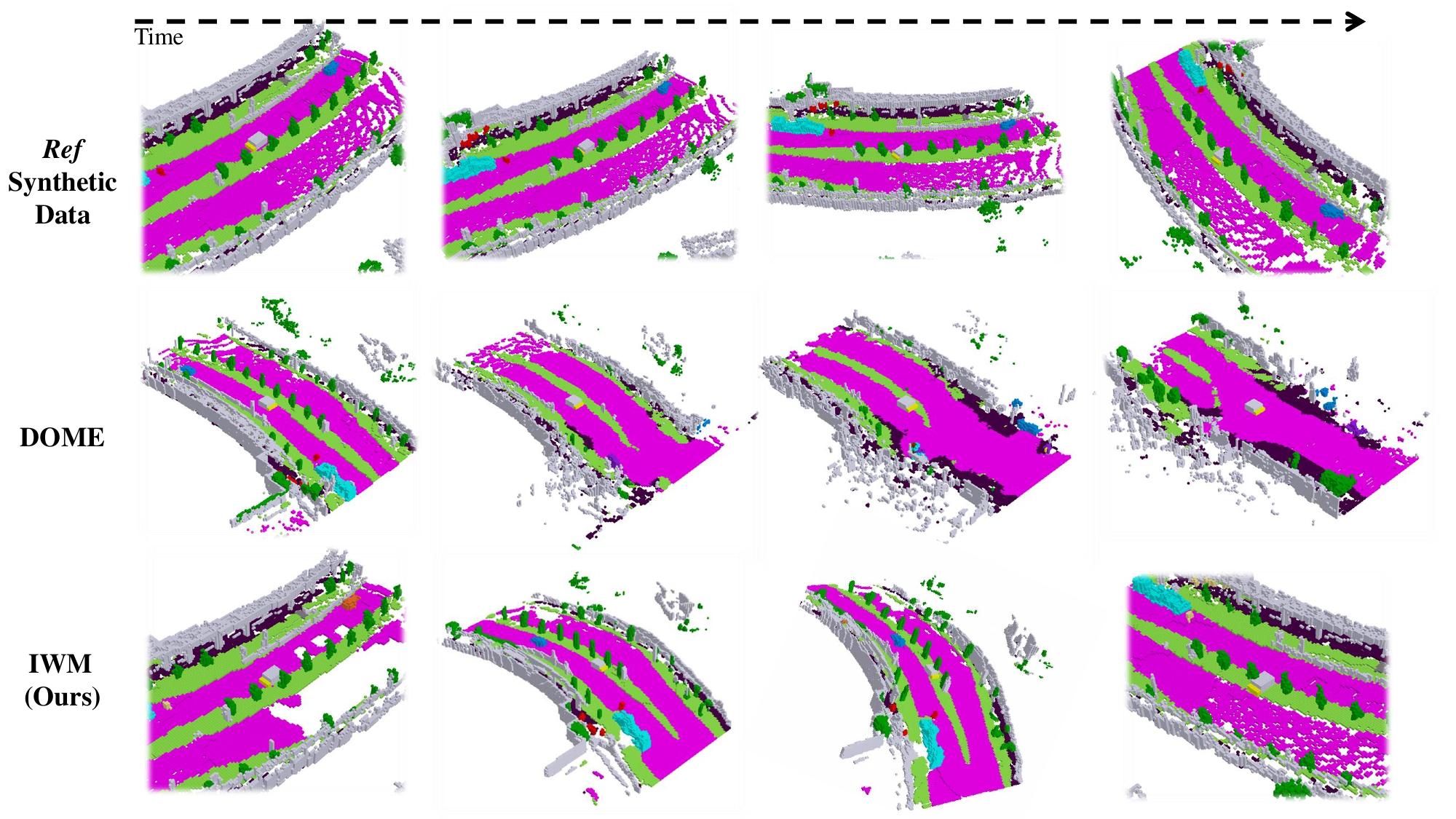}
\caption{Qualitative comparison of world models.\textbf{Top:} The reference synthetic data.  \textbf{Middle:} DOME\cite{gu2024dome}'s optimistic hallucination. \textbf{Bottom:} \ourmodel's faithful off-road prediction.}
\label{fig:qual_dream}
\vspace{-0.3cm}
\end{figure}

\noindent\textbf{Visualizing Failure Imagination.}
\cref{fig:qual_dream} provides a stark qualitative comparison. We feed an identical unsafe off-road trajectory into a standard world model~\cite{gu2024dome} and into \ourmodel. The standard model hallucinates a safe future, with the disappearing of the greenbelt in the middle of the road. In contrast, IWM faithfully predicts the off-road scene, generating an imagined future that provides a clear and correct penalty signal. More visualizations are incuded in the supplementary.

\noindent\textbf{Visualizing Policy Refinement}
In Figure~\ref{fig:qual_policy}, we show the behavior of a baseline agent versus our refined agent in a critical scenario. 
\begin{figure}[t]
\centering
\includegraphics[width=0.95\linewidth]{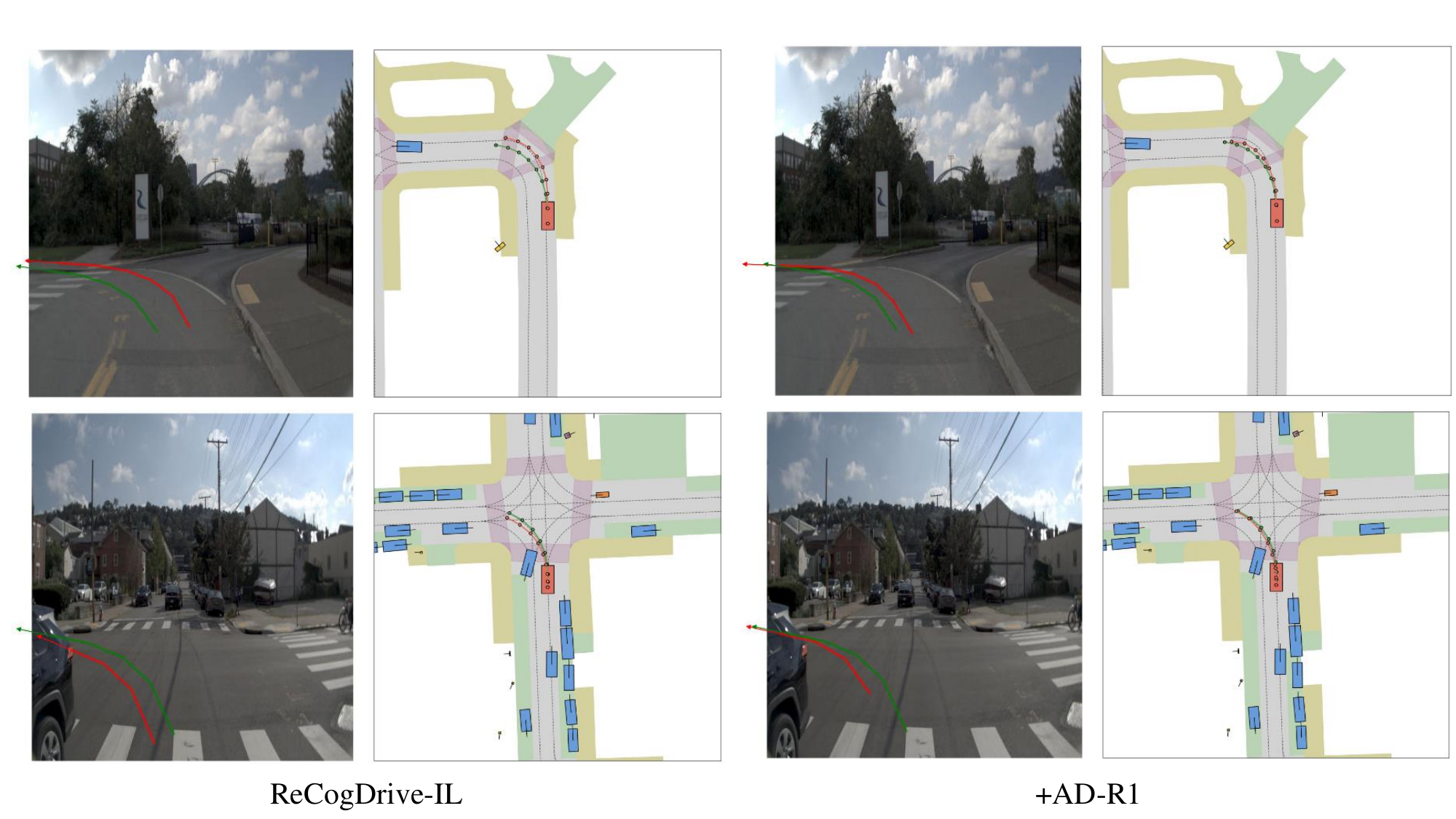}
\caption{Behavior of an agent with and without \textit{AD-R1} refinement. \textbf{Left:} The original agent's plan results in a collision or off-road. \textbf{Right:} Our refined agent safely avoids the hazard.}
\label{fig:qual_policy}
\end{figure}

\section{Conclusion}
Reinforcement learning for driving is stalled by a fundamental flaw: the ``optimistic bias" of world models that cannot predict failure. 
We solve this by introducing Counterfactual Synthesis to train an Impartial World Model that faithfully predicts the consequences of unsafe actions. 
As an internal critic in our \ourmodel~framework, this model enables agents to learn from imagined mistakes, drastically reducing safety violations in closed-loop simulations. 
We demonstrate that teaching a model to dream of danger is not a limitation, but a prerequisite for building truly safe and intelligent autonomous agents.
This work demonstrates that the path to robust autonomy is not to shield agents from failure, but to grant them the foresight to imagine it.

\clearpage\clearpage
{
    \small
    \bibliographystyle{ieeenat_fullname}
    \bibliography{main}

@String(AAAI = {AAAI})

@article{yan2025rlgf,
  title={Rlgf: Reinforcement learning with geometric feedback for autonomous driving video generation},
  author={Yan, Tianyi and Han, Wencheng and Zhou, Xia and Zhang, Xueyang and Zhan, Kun and Xu, Cheng-zhong and Shen, Jianbing},
  journal={arXiv preprint arXiv:2509.16500},
  year={2025}
}

@article{liang2025worldlens,
  title={WorldLens: Full-Spectrum Evaluations of Driving World Models in Real World},
  author={Liang, Ao and Kong, Lingdong and Yan, Tianyi and Liu, Hongsi and Yang, Wesley and Huang, Ziqi and Yin, Wei and Zuo, Jialong and Hu, Yixuan and Zhu, Dekai and others},
  journal={arXiv preprint arXiv:2512.10958},
  year={2025}
}

@inproceedings{hu2023uniad,
  title={Planning-oriented autonomous driving},
  author={Hu, Yihan and Yang, Jiazhi and Chen, Li and Li, Keyu and Sima, Chonghao and Zhu, Xizhou and Chai, Siqi and Du, Senyao and Lin, Tianwei and Wang, Wenhai and others},
  booktitle={Proceedings of the IEEE/CVF conference on computer vision and pattern recognition},
  pages={17853--17862},
  year={2023}
}

@inproceedings{jiang2023vad,
  title={Vad: Vectorized scene representation for efficient autonomous driving},
  author={Jiang, Bo and Chen, Shaoyu and Xu, Qing and Liao, Bencheng and Chen, Jiajie and Zhou, Helong and Zhang, Qian and Liu, Wenyu and Huang, Chang and Wang, Xinggang},
  booktitle={Proceedings of the IEEE/CVF International Conference on Computer Vision},
  pages={8340--8350},
  year={2023}
}

@article{chen2024vadv2,
  title={Vadv2: End-to-end vectorized autonomous driving via probabilistic planning},
  author={Chen, Shaoyu and Jiang, Bo and Gao, Hao and Liao, Bencheng and Xu, Qing and Zhang, Qian and Huang, Chang and Liu, Wenyu and Wang, Xinggang},
  journal={arXiv preprint arXiv:2402.13243},
  year={2024}
}

@article{kong20253d,
  title={3D and 4D World Modeling: A Survey},
  author={Kong, Lingdong and Yang, Wesley and Mei, Jianbiao and Liu, Youquan and Liang, Ao and Zhu, Dekai and Lu, Dongyue and Yin, Wei and Hu, Xiaotao and Jia, Mingkai and others},
  journal={arXiv preprint arXiv:2509.07996},
  year={2025}
}

@article{deng2025best3dscenerepresentation,
      title={What Is The Best 3D Scene Representation for Robotics? From Geometric to Foundation Models}, 
      author={Tianchen Deng and Yue Pan and Shenghai Yuan and Dong Li and Chen Wang and Mingrui Li and Long Chen and Lihua Xie and Danwei Wang and Jingchuan Wang and Javier Civera and Hesheng Wang and Weidong Chen},
      year={2025},
      journal={arXiv preprint arXiv:2512.03422}, 
}

@article{deng2025mcnslammultiagentcollaborativeneural,
      title={MCN-SLAM: Multi-Agent Collaborative Neural SLAM with Hybrid Implicit Neural Scene Representation}, 
      author={Tianchen Deng and Guole Shen and Xun Chen and Shenghai Yuan and Hongming Shen and Guohao Peng and Zhenyu Wu and Jingchuan Wang and Lihua Xie and Danwei Wang and Hesheng Wang and Weidong Chen},
      journal={arXiv preprint arXiv:2506.18678},
      year={2025},
}

@inproceedings{tang2025omnigen,
  title={OmniGen: Unified Multimodal Sensor Generation for Autonomous Driving},
  author={Tang, Tao and Ma, Enhui and Zhou, Xia and Wang, Letian and Yan, Tianyi and Zhang, Xueyang and Zhan, Kun and Jia, Peng and Lang, Xianpeng and Bian, Jia-Wang and others},
  booktitle={Proceedings of the 33rd ACM International Conference on Multimedia},
  pages={9365--9374},
  year={2025}
}

@article{yan2022plug,
  title={Plug-and-play pseudo label correction network for unsupervised person re-identification},
  author={Yan, Tianyi and Zhu, Kuan and Zhu, Guibo and Tang, Ming and Wang, Jinqiao and others},
  journal={arXiv preprint arXiv:2206.06607},
  year={2022}
}

@inproceedings{yan2025olidm,
  title={OLiDM: Object-aware LiDAR Diffusion Models for Autonomous Driving},
  author={Yan, Tianyi and Yin, Junbo and Lang, Xianpeng and Yang, Ruigang and Xu, Cheng-Zhong and Shen, Jianbing},
  booktitle={Proceedings of the AAAI Conference on Artificial Intelligence},
  volume={39},
  number={9},
  pages={9121--9129},
  year={2025}
}

@article{yan2024drivingsphere,
  title={DrivingSphere: Building a High-fidelity 4D World for Closed-loop Simulation},
  author={Yan, Tianyi and Wu, Dongming and Han, Wencheng and Jiang, Junpeng and Zhou, Xia and Zhan, Kun and Xu, Cheng-zhong and Shen, Jianbing},
  journal={arXiv preprint arXiv:2411.11252},
  year={2024}
}

@article{jia2023adriver,
  title={Adriver-i: A general world model for autonomous driving},
  author={Jia, Fan and Mao, Weixin and Liu, Yingfei and Zhao, Yucheng and Wen, Yuqing and Zhang, Chi and Zhang, Xiangyu and Wang, Tiancai},
  journal={arXiv preprint arXiv:2311.13549},
  year={2023}
}

@article{gao2024vista,
  title={Vista: A generalizable driving world model with high fidelity and versatile controllability},
  author={Gao, Shenyuan and Yang, Jiazhi and Chen, Li and Chitta, Kashyap and Qiu, Yihang and Geiger, Andreas and Zhang, Jun and Li, Hongyang},
  journal={Advances in Neural Information Processing Systems},
  volume={37},
  pages={91560--91596},
  year={2024}
}

@article{zhang2025epona,
  title={Epona: Autoregressive Diffusion World Model for Autonomous Driving},
  author={Zhang, Kaiwen and Tang, Zhenyu and Hu, Xiaotao and Pan, Xingang and Guo, Xiaoyang and Liu, Yuan and Huang, Jingwei and Yuan, Li and Zhang, Qian and Long, Xiao-Xiao and others},
  journal={arXiv preprint arXiv:2506.24113},
  year={2025}
}

@article{yang2024drivearena,
  title={Drivearena: A closed-loop generative simulation platform for autonomous driving},
  author={Yang, Xuemeng and Wen, Licheng and Ma, Yukai and Mei, Jianbiao and Li, Xin and Wei, Tiantian and Lei, Wenjie and Fu, Daocheng and Cai, Pinlong and Dou, Min and others},
  journal={arXiv preprint arXiv:2408.00415},
  year={2024}
}

@article{gao2024magicdrivedit,
  title={MagicDriveDiT: High-Resolution Long Video Generation for Autonomous Driving with Adaptive Control},
  author={Gao, Ruiyuan and Chen, Kai and Xiao, Bo and Hong, Lanqing and Li, Zhenguo and Xu, Qiang},
  journal={arXiv preprint arXiv:2411.13807},
  year={2024}
}

@article{wang2024occsora,
  title={Occsora: 4d occupancy generation models as world simulators for autonomous driving},
  author={Wang, Lening and Zheng, Wenzhao and Ren, Yilong and Jiang, Han and Cui, Zhiyong and Yu, Haiyang and Lu, Jiwen},
  journal={arXiv preprint arXiv:2405.20337},
  year={2024}
}

@inproceedings{zheng2024occworld,
  title={Occworld: Learning a 3d occupancy world model for autonomous driving},
  author={Zheng, Wenzhao and Chen, Weiliang and Huang, Yuanhui and Zhang, Borui and Duan, Yueqi and Lu, Jiwen},
  booktitle={European conference on computer vision},
  pages={55--72},
  year={2024},
  organization={Springer}
}

@article{bian2024dynamiccity,
  title={Dynamiccity: Large-scale lidar generation from dynamic scenes},
  author={Bian, Hengwei and Kong, Lingdong and Xie, Haozhe and Pan, Liang and Qiao, Yu and Liu, Ziwei},
  journal={arXiv e-prints},
  pages={arXiv--2410},
  year={2024}
}

@article{shi2025come,
  title={COME: Adding Scene-Centric Forecasting Control to Occupancy World Model},
  author={Shi, Yining and Jiang, Kun and Meng, Qiang and Wang, Ke and Wang, Jiabao and Sun, Wenchao and Wen, Tuopu and Yang, Mengmeng and Yang, Diange},
  journal={arXiv preprint arXiv:2506.13260},
  year={2025}
}

@article{law,
  title={Enhancing end-to-end autonomous driving with latent world model},
  author={Li, Yingyan and Fan, Lue and He, Jiawei and Wang, Yuqi and Chen, Yuntao and Zhang, Zhaoxiang and Tan, Tieniu},
  journal={arXiv preprint arXiv:2406.08481},
  year={2024}
}

@article{wote,
  title={End-to-end driving with online trajectory evaluation via bev world model},
  author={Li, Yingyan and Wang, Yuqi and Liu, Yang and He, Jiawei and Fan, Lue and Zhang, Zhaoxiang},
  journal={arXiv preprint arXiv:2504.01941},
  year={2025}
}

@inproceedings{zheng2025world4drive,
  title={World4Drive: End-to-end autonomous driving via intention-aware physical latent world model},
  author={Zheng, Yupeng and Yang, Pengxuan and Xing, Zebin and Zhang, Qichao and Zheng, Yuhang and Gao, Yinfeng and Li, Pengfei and Zhang, Teng and Xia, Zhongpu and Jia, Peng and others},
  booktitle={Proceedings of the IEEE/CVF International Conference on Computer Vision},
  pages={28632--28642},
  year={2025}
}

@article{gao2025rad,
  title={Rad: Training an end-to-end driving policy via large-scale 3dgs-based reinforcement learning},
  author={Gao, Hao and Chen, Shaoyu and Jiang, Bo and Liao, Bencheng and Shi, Yiang and Guo, Xiaoyang and Pu, Yuechuan and Yin, Haoran and Li, Xiangyu and Zhang, Xinbang and others},
  journal={arXiv preprint arXiv:2502.13144},
  year={2025}
}

@article{liu2025reinforced,
  title={Reinforced Refinement with Self-Aware Expansion for End-to-End Autonomous Driving},
  author={Liu, Haochen and Li, Tianyu and Yang, Haohan and Chen, Li and Wang, Caojun and Guo, Ke and Tian, Haochen and Li, Hongchen and Li, Hongyang and Lv, Chen},
  journal={arXiv preprint arXiv:2506.09800},
  year={2025}
}

@article{dauner2024navsim,
  title={Navsim: Data-driven non-reactive autonomous vehicle simulation and benchmarking},
  author={Dauner, Daniel and Hallgarten, Marcel and Li, Tianyu and Weng, Xinshuo and Huang, Zhiyu and Yang, Zetong and Li, Hongyang and Gilitschenski, Igor and Ivanovic, Boris and Pavone, Marco and others},
  journal={Advances in Neural Information Processing Systems},
  volume={37},
  pages={28706--28719},
  year={2024}
}

@inproceedings{dosovitskiy2017carla,
  title={CARLA: An open urban driving simulator},
  author={Dosovitskiy, Alexey and Ros, German and Codevilla, Felipe and Lopez, Antonio and Koltun, Vladlen},
  booktitle={Conference on robot learning},
  pages={1--16},
  year={2017},
  organization={PMLR}
}

@article{guo2025deepseek,
  title={Deepseek-r1: Incentivizing reasoning capability in llms via reinforcement learning},
  author={Guo, Daya and Yang, Dejian and Zhang, Haowei and Song, Junxiao and Zhang, Ruoyu and Xu, Runxin and Zhu, Qihao and Ma, Shirong and Wang, Peiyi and Bi, Xiao and others},
  journal={arXiv preprint arXiv:2501.12948},
  year={2025}
}

@article{lu2025vla-rl,
  title={Vla-rl: Towards masterful and general robotic manipulation with scalable reinforcement learning},
  author={Lu, Guanxing and Guo, Wenkai and Zhang, Chubin and Zhou, Yuheng and Jiang, Haonan and Gao, Zifeng and Tang, Yansong and Wang, Ziwei},
  journal={arXiv preprint arXiv:2505.18719},
  year={2025}
}

@article{gao2023magicdrive,
  title={Magicdrive: Street view generation with diverse 3d geometry control},
  author={Gao, Ruiyuan and Chen, Kai and Xie, Enze and Hong, Lanqing and Li, Zhenguo and Yeung, Dit-Yan and Xu, Qiang},
  journal={arXiv preprint arXiv:2310.02601},
  year={2023}
}

@inproceedings{chen2025drivinggpt,
  title={Drivinggpt: Unifying driving world modeling and planning with multi-modal autoregressive transformers},
  author={Chen, Yuntao and Wang, Yuqi and Zhang, Zhaoxiang},
  booktitle={Proceedings of the IEEE/CVF International Conference on Computer Vision},
  pages={26890--26900},
  year={2025}
}

@article{chitta2022transfuser,
  title={Transfuser: Imitation with transformer-based sensor fusion for autonomous driving},
  author={Chitta, Kashyap and Prakash, Aditya and Jaeger, Bernhard and Yu, Zehao and Renz, Katrin and Geiger, Andreas},
  journal={IEEE transactions on pattern analysis and machine intelligence},
  volume={45},
  number={11},
  pages={12878--12895},
  year={2022},
  publisher={IEEE}
}

@article{li2024hydra,
  title={Hydra-mdp: End-to-end multimodal planning with multi-target hydra-distillation},
  author={Li, Zhenxin and Li, Kailin and Wang, Shihao and Lan, Shiyi and Yu, Zhiding and Ji, Yishen and Li, Zhiqi and Zhu, Ziyue and Kautz, Jan and Wu, Zuxuan and others},
  journal={arXiv preprint arXiv:2406.06978},
  year={2024}
}

@inproceedings{liao2025i2,
  title={I2-World: Intra-inter tokenization for efficient dynamic 4D scene forecasting},
  author={Liao, Zhimin and Wei, Ping and Zhang, Ruijie and Chen, Shuaijia and Wang, Haoxuan and Ren, Ziyang},
  booktitle={Proceedings of the IEEE/CVF International Conference on Computer Vision},
  pages={25810--25819},
  year={2025}
}

@article{yang2025raw2drive,
  title={Raw2Drive: Reinforcement learning with aligned world models for end-to-end autonomous driving (in carla v2)},
  author={Yang, Zhenjie and Jia, Xiaosong and Li, Qifeng and Yang, Xue and Yao, Maoqing and Yan, Junchi},
  journal={arXiv preprint arXiv:2505.16394},
  year={2025}
}

@inproceedings{li2024think2drive,
  title={Think2drive: Efficient reinforcement learning by thinking with latent world model for autonomous driving (in carla-v2)},
  author={Li, Qifeng and Jia, Xiaosong and Wang, Shaobo and Yan, Junchi},
  booktitle={European Conference on Computer Vision},
  pages={142--158},
  year={2024},
  organization={Springer}
}

@inproceedings{sun2020waymo,
  title={Scalability in perception for autonomous driving: Waymo open dataset},
  author={Sun, Pei and Kretzschmar, Henrik and Dotiwalla, Xerxes and Chouard, Aurelien and Patnaik, Vijaysai and Tsui, Paul and Guo, James and Zhou, Yin and Chai, Yuning and Caine, Benjamin and others},
  booktitle={Proceedings of the IEEE/CVF conference on computer vision and pattern recognition},
  pages={2446--2454},
  year={2020}
}

@article{tian2023occ3d,
  title={Occ3d: A large-scale 3d occupancy prediction benchmark for autonomous driving},
  author={Tian, Xiaoyu and Jiang, Tao and Yun, Longfei and Mao, Yucheng and Yang, Huitong and Wang, Yue and Wang, Yilun and Zhao, Hang},
  journal={Advances in Neural Information Processing Systems},
  volume={36},
  pages={64318--64330},
  year={2023}
}

@article{schubert2017dbscan,
  title={DBSCAN revisited, revisited: why and how you should (still) use DBSCAN},
  author={Schubert, Erich and Sander, J{\"o}rg and Ester, Martin and Kriegel, Hans Peter and Xu, Xiaowei},
  journal={ACM Transactions on Database Systems (TODS)},
  volume={42},
  number={3},
  pages={1--21},
  year={2017},
  publisher={Acm New York, NY, USA}
}

@article{shang2025drivedpo,
  title={DriveDPO: Policy Learning via Safety DPO For End-to-End Autonomous Driving},
  author={Shang, Shuyao and Chen, Yuntao and Wang, Yuqi and Li, Yingyan and Zhang, Zhaoxiang},
  journal={arXiv preprint arXiv:2509.17940},
  year={2025}
}

@article{li2025trajhf,
  title={Finetuning generative trajectory model with reinforcement learning from human feedback},
  author={Li, Derun and Ren, Jianwei and Wang, Yue and Wen, Xin and Li, Pengxiang and Xu, Leimeng and Zhan, Kun and Xia, Zhongpu and Jia, Peng and Lang, Xianpeng and others},
  journal={arXiv preprint arXiv:2503.10434},
  year={2025}
}

@inproceedings{wang2024drivedreamer,
  title={DriveDreamer: Towards Real-World-Drive World Models for Autonomous Driving},
  author={Wang, Xiaofeng and Zhu, Zheng and Huang, Guan and Chen, Xinze and Zhu, Jiagang and Lu, Jiwen},
  booktitle={European Conference on Computer Vision},
  pages={55--72},
  year={2024},
  organization={Springer}
}

@inproceedings{zhao2025drivedreamer2,
  title={Drivedreamer-2: Llm-enhanced world models for diverse driving video generation},
  author={Zhao, Guosheng and Wang, Xiaofeng and Zhu, Zheng and Chen, Xinze and Huang, Guan and Bao, Xiaoyi and Wang, Xingang},
  booktitle={Proceedings of the AAAI Conference on Artificial Intelligence},
  volume={39},
  number={10},
  pages={10412--10420},
  year={2025}
}

@article{ouyang2022rlhf,
  title={Training language models to follow instructions with human feedback},
  author={Ouyang, Long and Wu, Jeffrey and Jiang, Xu and Almeida, Diogo and Wainwright, Carroll and Mishkin, Pamela and Zhang, Chong and Agarwal, Sandhini and Slama, Katarina and Ray, Alex and others},
  journal={Advances in neural information processing systems},
  volume={35},
  pages={27730--27744},
  year={2022}
}

@inproceedings{caesar2020nuscenes,
  title={nuscenes: A multimodal dataset for autonomous driving},
  author={Caesar, Holger and Bankiti, Varun and Lang, Alex H and Vora, Sourabh and Liong, Venice Erin and Xu, Qiang and Krishnan, Anush and Pan, Yu and Baldan, Giancarlo and Beijbom, Oscar},
  booktitle={Proceedings of the IEEE/CVF conference on computer vision and pattern recognition},
  pages={11621--11631},
  year={2020}
}

@article{li2025recogdrive,
  title={Recogdrive: A reinforced cognitive framework for end-to-end autonomous driving},
  author={Li, Yongkang and Xiong, Kaixin and Guo, Xiangyu and Li, Fang and Yan, Sixu and Xu, Gangwei and Zhou, Lijun and Chen, Long and Sun, Haiyang and Wang, Bing and others},
  journal={arXiv preprint arXiv:2506.08052},
  year={2025}
}

@article{renz2024carllava,
  title={Carllava: Vision language models for camera-only closed-loop driving},
  author={Renz, Katrin and Chen, Long and Marcu, Ana-Maria and H{\"u}nermann, Jan and Hanotte, Benoit and Karnsund, Alice and Shotton, Jamie and Arani, Elahe and Sinavski, Oleg},
  journal={arXiv preprint arXiv:2406.10165},
  year={2024}
}

@inproceedings{liao2025diffusiondrive,
  title={Diffusiondrive: Truncated diffusion model for end-to-end autonomous driving},
  author={Liao, Bencheng and Chen, Shaoyu and Yin, Haoran and Jiang, Bo and Wang, Cheng and Yan, Sixu and Zhang, Xinbang and Li, Xiangyu and Zhang, Ying and Zhang, Qian and others},
  booktitle={Proceedings of the Computer Vision and Pattern Recognition Conference},
  pages={12037--12047},
  year={2025}
}

@inproceedings{xing2025goalflow,
  title={Goalflow: Goal-driven flow matching for multimodal trajectories generation in end-to-end autonomous driving},
  author={Xing, Zebin and Zhang, Xingyu and Hu, Yang and Jiang, Bo and He, Tong and Zhang, Qian and Long, Xiaoxiao and Yin, Wei},
  booktitle={Proceedings of the Computer Vision and Pattern Recognition Conference},
  pages={1602--1611},
  year={2025}
}

@inproceedings{liao2025i2world,
  title={I2-World: Intra-inter tokenization for efficient dynamic 4D scene forecasting},
  author={Liao, Zhimin and Wei, Ping and Zhang, Ruijie and Chen, Shuaijia and Wang, Haoxuan and Ren, Ziyang},
  booktitle={Proceedings of the IEEE/CVF International Conference on Computer Vision},
  pages={25810--25819},
  year={2025}
}

@article{gu2024dome,
  title={Dome: Taming diffusion model into high-fidelity controllable occupancy world model},
  author={Gu, Songen and Yin, Wei and Jin, Bu and Guo, Xiaoyang and Wang, Junming and Li, Haodong and Zhang, Qian and Long, Xiaoxiao},
  journal={arXiv preprint arXiv:2410.10429},
  year={2024}
}
}

\onecolumn
\setcounter{section}{0}
\renewcommand{\thesection}{\Alph{section}}
\maketitlesupplementary
\onecolumn
\setcounter{page}{1}
\begin{center}
   \Large \textbf{AD-R1: Closed-Loop Reinforcement Learning for End-to-End Autonomous Driving with Impartial World Models}
\end{center}
\begin{center}
   \Large Supplementary Material
\end{center}
\section*{Overview}
This supplementary material provides additional details to support the claims and ensure the reproducibility of the main paper. The content is organized as follows:
\begin{itemize}
    \item \textbf{Sec. \ref{sec:implementation}}: Detailed network architecture, hyperparameters for the Impartial World Model (IWM), and the RL training configuration.
    \item \textbf{Sec. \ref{sec:dataset}}: Elaboration on the nuScenes-CF dataset and the mathematical formulation of Counterfactual Synthesis.
    \item \textbf{Sec. \ref{sec:setup}}: Strict mathematical definitions for the Risk Foreseeing Benchmark (RFB).
    \item \textbf{Sec. \ref{sec:add_exp}}: Additional quantitative results focusing on the sensitivity analysis of synthetic data ratios.
    \item \textbf{Sec. \ref{sec:viz}}: Qualitative analysis and description of the attached demo.
\end{itemize}

\section{Implementation Details}
\label{sec:implementation}

\subsection{Impartial World Model (IWM) Architecture}
Our IWM consists of a VQ-VAE-based tokenizer and a Transformer-based autoregressive forecaster based on I$^2$-World~\cite{liao2025i2}.

\noindent\textbf{4D Scene Tokenizer.} 
We voxelize the local scene into a grid of spatial resolution $200 \times 200$ and height resolution $16$. The physical range covered is $[-40\text{m}, 40\text{m}]$ in $X, Y$ and $[-1\text{m}, 5.4\text{m}]$ in $Z$, resulting in a voxel size of $0.4\text{m} \times 0.4\text{m} \times 0.4\text{m}$, following Occ-3D~\cite{tian2023occ3d}.
\begin{itemize}
    \item \textbf{Encoder/Decoder:} We utilize a 3D-VAE architecture with 4 downsampling stages. 
    \item \textbf{Vector Quantization:} We use a codebook of size $N=512$ with an embedding dimension of $D=128$.
\end{itemize}

\noindent\textbf{4D Forecaster.}
The forecaster is a encoder-decoder Transformer tailored for 4D occupancy generation.
\begin{itemize}
    \item \textbf{Structure:} 24 Attention Layers, 16 Attention Heads, Hidden Dimension $d_{model}=1024$.
    \item \textbf{Context:} The model takes 2 seconds of historical context (4 frames at 2Hz) and predicts 3 seconds into the future (6 frames).
\end{itemize}
\noindent\textbf{Training Details.} We use AdamW optimizer with $1\times10^{-3}$ as the base learning rate, 256 as the batch size. The Tokenizer is trained for 30 epochs while the Forecaster is trained for 50 epochs. All training is performed on $8 \times $NVIDIA H20 GPUs.

\subsection{RL Training Configuration (AD-R1)}
We employ the Group Relative Policy Optimization (GRPO) algorithm adapted for continuous trajectory refinement.

\noindent\textbf{Training Setup.}
\begin{itemize}
    \item \textbf{Policy Network:} We fine-tune the denoising head of the pre-trained DiffusionDrive \cite{liao2025diffusiondrive} and ReCogDrive~\cite{li2025recogdrive}.
    \item \textbf{Group Sampling ($G$):} Due to the computational cost of 4D world model rollouts, we set the group size $G=64$ and $G=8$ for DiffusionDrive and ReCogDrive, respectively. This provides a sufficient baseline variance for advantage estimation while fitting within GPU memory constraints.
    \item \textbf{Optimization:} We use the AdamW optimizer with a learning rate of $1 \times 10^{-5}$ and $4 \times 10^{-5}$ for DiffusionDrive and ReCogDrive.
    \item \textbf{Hardware:} Training is performed on 8 $\times$ NVIDIA H20 GPUs. The refinement stage takes approximately 24 hours for 10 epochs on the navsim~\cite{dauner2024navsim} training set.
\end{itemize}

\subsection{Reward Function Details}
The reward shaping is critical for guiding the agent towards safety without compromising progress. Table \ref{tab:reward_weights} details the specific weights used in Eq. (18) of the main paper.

\section{The nuScenes-CF Dataset}
\label{sec:dataset}

\subsection{Kinematic Trajectory Generation}
To generate the unsafe ego-trajectory $\tilde{\mathcal{T}}_{ego}$ for Counterfactual Synthesis, we employ a Blending Kinematic Model. Let $\mathbf{p}_t, \mathbf{v}_t$ be the vehicle's current state and $\mathbf{p}_{target}$ be the designated collision point (e.g., the center of another vehicle or a wall).

The synthesized position $\mathbf{p}_{t+k}$ at future timestep $k$ is computed as:
\begin{equation}
    \mathbf{p}_{t+k} = (1 - \gamma_k) \cdot \mathbf{p}_{inertial}^{(k)} + \gamma_k \cdot \mathbf{p}_{intercept}^{(k)}
\end{equation}
where:
\begin{itemize}
    \item $\mathbf{p}_{inertial}^{(k)}$ represents the position if the vehicle continued with its historical momentum.
    \item $\mathbf{p}_{intercept}^{(k)}$ is the position required to linearly intercept the target $\mathbf{p}_{target}$.
    \item $\gamma_k \in [0, 1]$ is a time-varying blending coefficient, defined as $\gamma_k = (k/K)^2$. The quadratic growth ensures a smooth deviation that mimics a loss of control or a gradual drift, rather than an unrealistic sharp turn.
\end{itemize}

\subsection{Dataset Statistics and Sampling Strategy}
We constructed a dedicated pool of counterfactual scenarios consisting of \textbf{1,800 synthetic clips} derived from the nuScenes training set. For evaluation, we curated a separate set of \textbf{450 clips} to serve as the Risk Foreseeing Benchmark (RFB). 

\noindent\textbf{Training Sampling Strategy.}
It is important to note that the "80:20 ratio" refers to the data distribution seen by the model during training, not the absolute size of the datasets. We employ a \textbf{weighted sampling strategy}: in each training iteration, 80\% of the samples are drawn from the original real-world nuScenes dataset (safe expert data), and 20\% are sampled from our synthetic counterfactual pool (unsafe failure data). This ensures the model maintains strong priors on realistic scene dynamics while receiving sufficient exposure to rare failure modes.

The distribution of failure types within the synthetic pool is balanced as follows:
\begin{itemize}
    \item \textbf{Dynamic Collisions (40\%):} Inter-vehicle collisions and vehicle-pedestrian accidents.
    \item \textbf{Static Collisions (30\%):} Collisions with barriers, traffic cones, and buildings.
    \item \textbf{Off-Road/Sidewalk (30\%):} Excursions into non-drivable zones.
\end{itemize}
\subsection{Counterfactual Example}
\begin{figure}[t]
    \centering
    \includegraphics[width=0.99\linewidth]{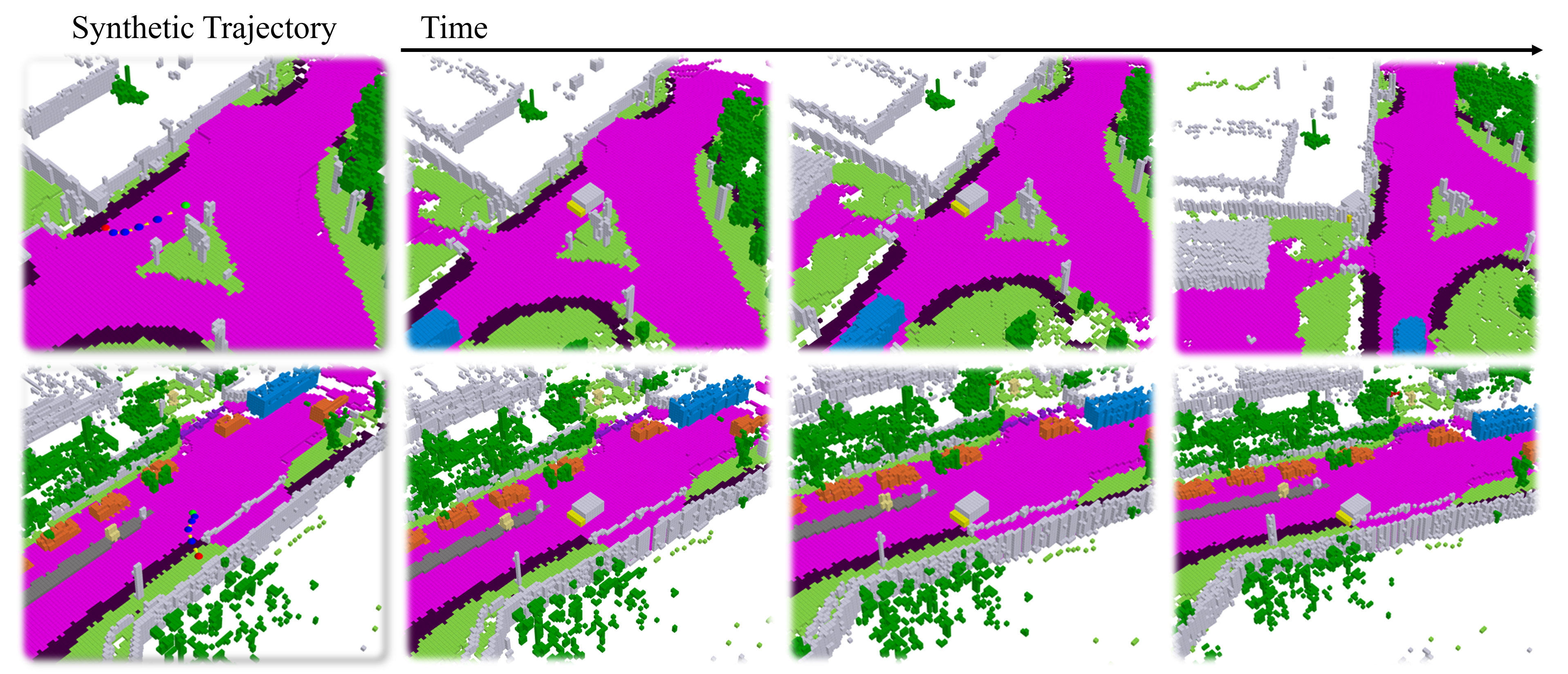}
    \caption{Examples of synthetic unsafe trajectories for the Counterfactual Data (Off-road). Red dot indicates the end of the trajectory while the green one indicates the start point.}
    \label{fig:data_offroad}
\end{figure}

\begin{figure}[t]
    \centering
    \includegraphics[width=0.99\linewidth]{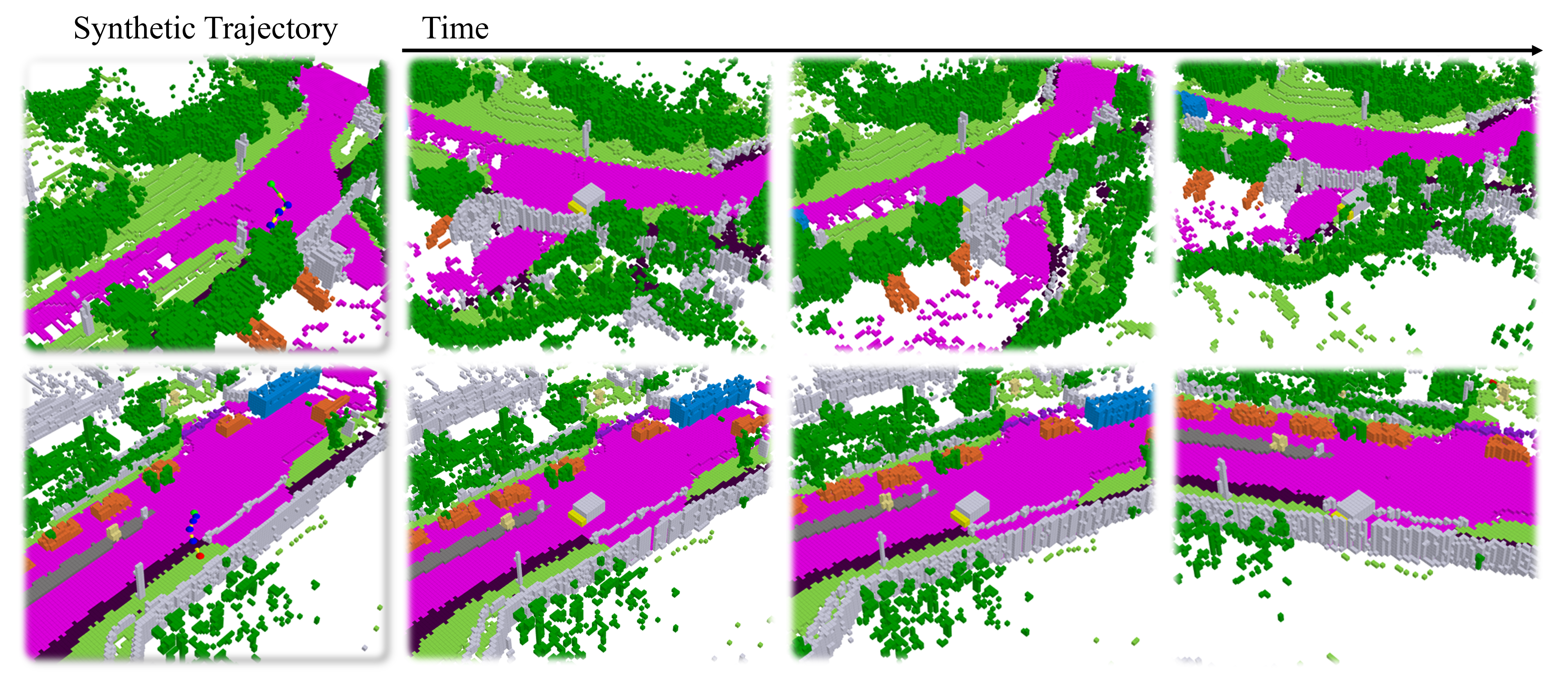}
    \caption{Examples of synthetic unsafe trajectories for the Counterfactual Data (Penetration). Red dot indicates the end of the trajectory while the green one indicates the start point.}
    \label{fig:data_pen}
\end{figure}

\begin{figure}[t]
    \centering
    \includegraphics[width=0.99\linewidth]{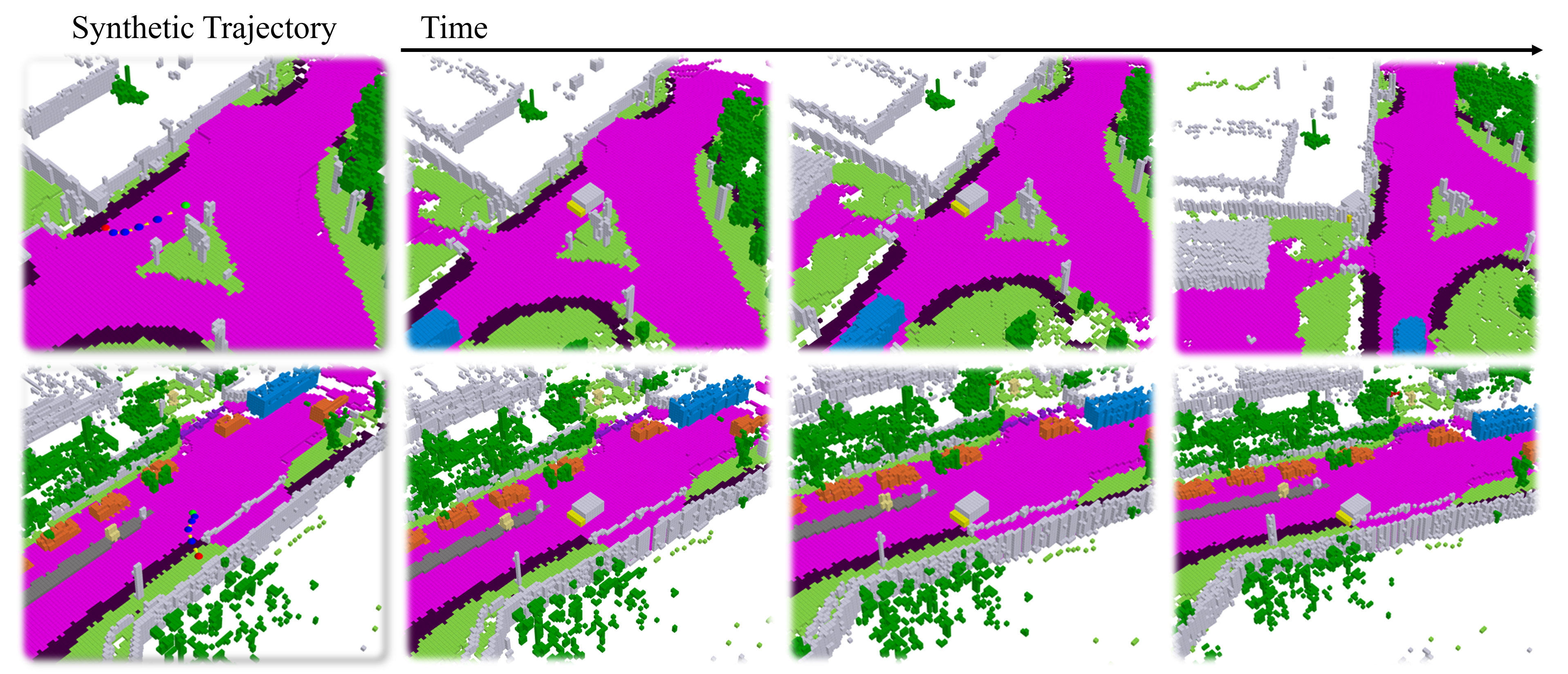}
    \caption{Examples of synthetic unsafe trajectories for the Counterfactual Data (Collision). Red dot indicates the end of the trajectory while the green one indicates the start point.}
    \label{fig:data_collision}
\end{figure}

We visualize examples about synthetic counterfactual occupancy data to show \textbf{Penetration} (\cref{fig:data_pen}),  \textbf{Off-road} (\cref{fig:data_offroad}) and \textbf{Collision} (\cref{fig:data_collision}).

\section{Experimental Setup and Benchmarks}
\label{sec:setup}

\subsection{Risk Foreseeing Benchmark (RFB)}
To quantify the core contribution of our AD-R1 for overcoming the optimistic bias, we evaluate world models on our proposed \textbf{Risk Foreseeing Benchmark (RFB)}. The RFB consists of a curated set of \textbf{450 challenging scenarios} (clips) with ground-truth unsafe ego-trajectories spanning 3 seconds. Unlike standard benchmarks that evaluate on safe expert driving, RFB strictly tests the model's ability to predict negative outcomes.

We measure performance using the following three key metrics:

\begin{enumerate}
    \item \textbf{Global Scene Fidelity (\textit{G}-IoU):} 
    This metric calculates the mean Intersection-over-Union (mIoU) across the entire scene (including static background and dynamic agents). It serves as a sanity check to penalize unrelated hallucinations and ensure the overall scene structure remains consistent.
    \begin{equation}
        \text{\textit{G}-IoU} = \frac{1}{|C|} \sum_{c \in C} \frac{|\hat{O}_c \cap O_c|}{|\hat{O}_c \cup O_c|}
    \end{equation}
    where $C$ represents the semantic classes in the scene.

    \item \textbf{Failure IoU (\textit{f}-IoU):} 
    To strictly measure the prediction of danger, we define \textit{f}-IoU. This metric focuses on the \textbf{critical failure regions} (e.g., collision zones surrounding the ego car). We define a local volume $V_{crit}$ centered around the ego-vehicle's future trajectory. \textit{f}-IoU calculates the IoU of occupied voxels specifically within this volume:
    \begin{equation}
        \text{\textit{f}-IoU} = \frac{|\hat{O}_{occ} \cap O_{occ} \cap V_{crit}|}{|\hat{O}_{occ} \cup O_{occ} \cap V_{crit}|}
    \end{equation}
    A high \textit{f}-IoU indicates the model correctly predicts that "something is there" (e.g., a wall or another car) directly in the ego-vehicle's path, rather than hallucinating empty space.

    \item \textbf{Dynamic Agent Fidelity (DAF):} 
    Previous optimistic models tend to make other agents vanish ("ghosting") to avoid collisions. DAF measures the average Instance-IoU of other dynamic agents (Vehicles, Pedestrians) to ensure that risk-inducing actors are preserved in the prediction.
    \begin{equation}
        \text{DAF} = \frac{1}{N_{obj}} \sum_{i=1}^{N_{obj}} \text{IoU}(\hat{M}_i, M_i)
    \end{equation}
    where $M_i$ is the occupancy mask of the $i$-th dynamic agent.
\end{enumerate}

\subsection{Driving Policy Evaluation}
On the NavSim benchmark, we evaluate the final agent's performance using safety and planning metrics. For planning quality, we use the \textbf{PDSM}~\cite{dauner2024navsim}, which provides a holistic score for driving quality, weighting progress, comfort, and safety.

\subsection{Baselines and Comparison Schemes}
We conduct a comprehensive comparison to validate both the world model and the downstream policy refinement.

\noindent\textbf{1. For World Model Evaluation:} 
We compare our \textbf{Impartial World Model (IWM)} against strong baseline world models trained only on safe data to demonstrate the impact of optimistic bias:
\begin{itemize}
    \item \textbf{I$^2$-World}~\cite{liao2025i2world}: A state-of-the-art 4D occupancy world model using VQ-VAE and autoregressive transformers.
    \item \textbf{DOME}~\cite{gu2024dome}: A diffusion-based occupancy world model.
\end{itemize}

\noindent\textbf{2. For Policy Refinement:} 
We apply our plug-and-play framework, powered by IWM, to two state-of-the-art, publicly available, pre-trained end-to-end driving agents. We compare the original performance against our refined versions:
\begin{itemize}
    \item \textbf{DiffusionDrive}~\cite{liao2025diffusiondrive}: A diffusion-based planning policy.
    \item \textbf{ReCogDrive}~\cite{li2025recogdrive}: A VLA-based agent integrating vision-language features.
\end{itemize}

\begin{table}[t]
    \centering
    \small
    \caption{\textbf{Reward Coefficients Configuration.} Negative weights indicate penalties.}
    \label{tab:reward_weights} 
    \begin{tabular}{l c c p{4cm}}
        \toprule
        \textbf{Component} & \textbf{Symbol} & \textbf{Weight ($w$)} & \textbf{Description} \\
        \midrule
        \multicolumn{4}{l}{\textit{Safety-Critical (Heavily Penalized)}} \\
        Collision & $w_{coll}$ & $-20.0$ & Penalty if VC-IoU $> 0$. \\
        Off-Road & $w_{off}$ & $-10.0$ & Penalty for non-drivable areas. \\
        Clearance & $w_{clr}$ & $-15.0$ & Vertical height violation. \\
        \midrule
        \multicolumn{4}{l}{\textit{Comfort \& Task}} \\
        Stability & $w_{stab}$ & $-2.0$ & Variance of Z-axis under ego. \\
        Progress & $w_{prog}$ & $+1.0$ & Reward for distance traveled. \\
        Velocity & $w_{vel}$ & $+0.5$ & Alignment with speed limit. \\
        \bottomrule
    \end{tabular}
\end{table}
\section{Additional Experimental Results}
\label{sec:add_exp}

\subsection{Ablation: Synthetic Data Ratio}
We investigate the impact of the ratio of synthetic counterfactual data mixed into the training set. Table \ref{tab:data_ratio} shows that a balance is crucial.

\begin{table}[htbp]
    \centering
    \caption{\textbf{Effect of Synthetic Data Ratio.} We find that an 80:20 mix offers the best trade-off between scene fidelity (G-IoU) and safety awareness (f-IoU).}
    \label{tab:data_ratio}
    \begin{tabular}{c c c c}
        \toprule
        \textbf{Real : Syn Ratio} & \textbf{G-IoU} $\uparrow$ & \textbf{f-IoU} $\uparrow$ & \textbf{PDMS} $\uparrow$ \\
        \midrule
        100 : 0 (Baseline) & 21.01 & 14.21 & 85.3 \\
        90 : 10 & 35.40 & 38.12 & 88.2 \\
        \textbf{80 : 20 (Ours)} & \textbf{40.21} & 45.91 & \textbf{89.8} \\
        50 : 50 & 38.50 & \textbf{46.10} & 87.4 \\
        \bottomrule
    \end{tabular}
\end{table}
While increasing synthetic data to 50\% slightly improves f-IoU (collision prediction), it degrades the general scene generation quality (G-IoU) because the synthetic physics are simplified approximations. The 80:20 ratio provides the optimal balance.

\subsection{Ablation: Reward Function Components}
To validate the contribution of each component in our multi-faceted reward function $R_{total}$, we conduct an ablation study based on DiffusionDrive by selectively removing specific reward terms during the RL training phase. Table~\ref{tab:reward_ablation} summarizes the impact on the final policy performance.

\begin{table}[t]
    \centering
    \caption{\textbf{Ablation on Reward Components.} Each component plays a distinct role: Safety penalties are crucial for collision avoidance, Progress rewards prevent the "frozen robot" problem, and Comfort rewards ensure smooth 3D trajectory planning.}
    \label{tab:reward_ablation}
    \begin{tabular}{l c  c}
        \toprule
        \textbf{Configuration} & \textbf{Coll. Rate} $\downarrow$  & \textbf{PDMS} $\uparrow$ \\
        \midrule
        DiffusionDrive & 1.8\%  & 88.1 \\
        \textbf{Full AD-R1 (Ours)} & \textbf{1.6\%}  & \textbf{89.8} \\
        \midrule
        w/o Safety ($w_{coll}, w_{off}, w_{clr} = 0$) & 1.8\%  & 88.5 \\
        w/o Comfort ($w_{stab} = 0$) & 1.8\% &  87.6 \\
        w/o Progress ($w_{prog}, w_{vel} = 0$) & 2.0\%  & 76.8 \\
        \bottomrule
    \end{tabular}
\end{table}

Impact of Safety Penalties: When safety penalties are removed (Row 3), the Collision Rate reverts to the baseline level of 1.8\%. While the absolute difference (0.2\%) appears small, it represents a meaningful improvement given that the baseline policy is already highly optimized. More importantly, our Impartial World Model provides a mechanism to penalize subtle unsafe behaviors that heuristics miss.
\textbf{Impact of Comfort:} Removing the 3D stability reward (Row 4) results in a PDMS of 87.6, which is even lower than the original baseline (88.1). This suggests that without comfort constraints, the RL process may optimize for safety or speed at the expense of ride smoothness, generating jerky trajectories that degrade the overall driving quality.
\textbf{Impact of Progress:} Removing task-oriented rewards (Row 5) causes a catastrophic drop in PDMS to 76.8. Interestingly, the collision rate slightly increases to 2.0\%, implying that an agent lacking goal-directed behavior may behave unpredictably or fail to exit dangerous zones efficiently, thereby increasing risk.

\section{Qualitative Visualization}
\label{sec:viz}

In Figure~\ref{fig:qual_policy}, we show the behavior of a baseline agent versus our refined agent in a critical scenario. The original agent, failing to anticipate a sudden cut-in, generates a plan that leads to a collision. Our refined agent, having learned from countless imagined failures, exhibits more defensive behavior, correctly predicting the risk and executing a safe braking maneuver.
\begin{figure}[t]
\centering
\includegraphics[width=0.95\linewidth]{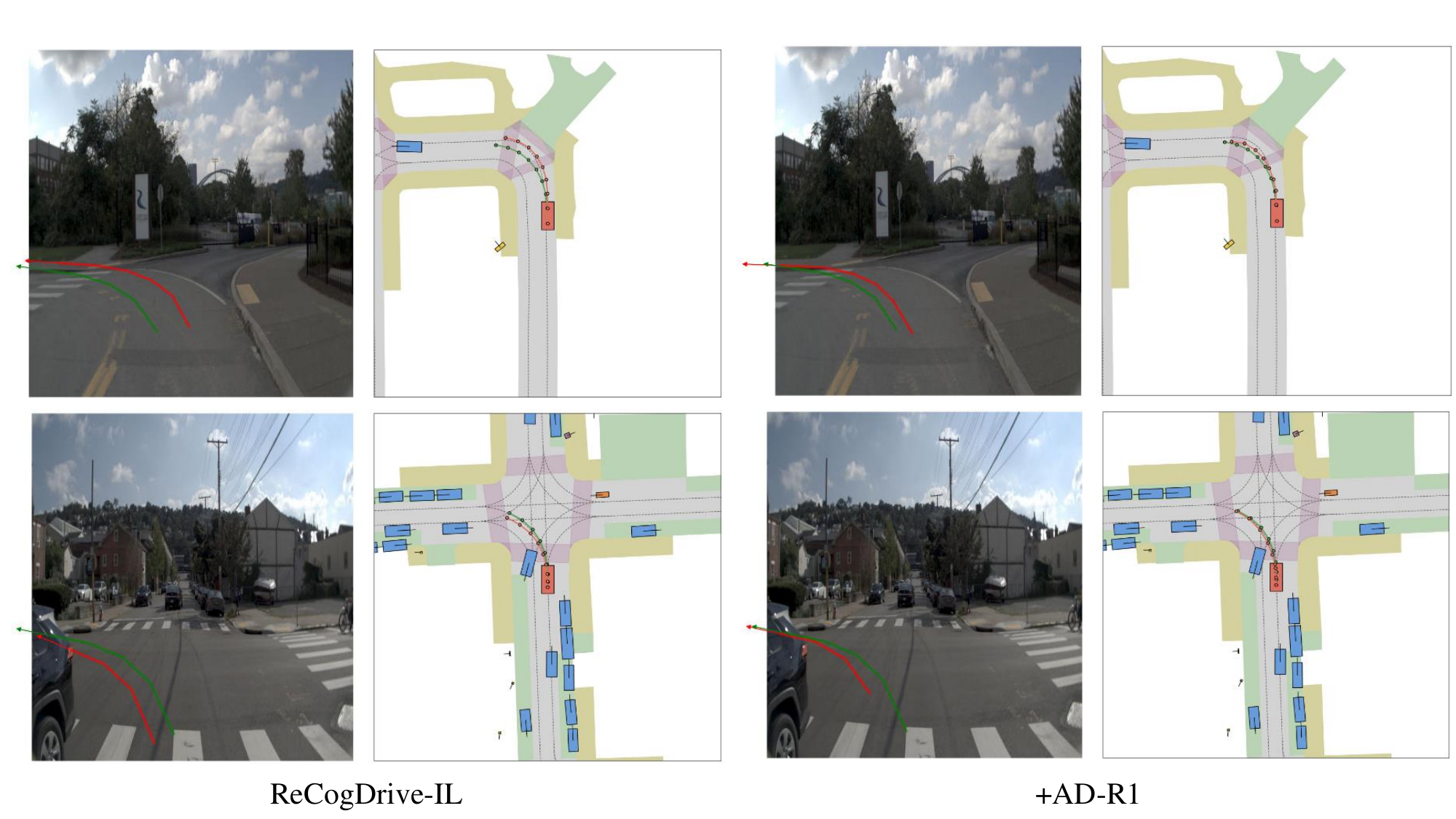}
\caption{Behavior of an agent with and without \textit{AD-R1} refinement. \textbf{Left:} The original agent's plan results in a collision or off-road. \textbf{Right:} Our refined agent safely avoids the hazard.}
\label{fig:qual_policy}
\end{figure}

\end{document}